\documentclass{article}

\usepackage{arxiv}
\usepackage[utf8]{inputenc}
\usepackage[T1]{fontenc}
\usepackage{hyperref}
\usepackage{url}
\usepackage{booktabs}
\usepackage{amsfonts}
\usepackage{amsmath}
\usepackage{nicefrac}
\usepackage{microtype}
\usepackage{cleveref}
\usepackage{graphicx}
\usepackage[numbers]{natbib}
\usepackage{doi}
\usepackage{listings}
\usepackage{xcolor}

\usepackage{array}
\usepackage{float}
\usepackage{makecell}
\usepackage{tabularx}
\usepackage{ragged2e}
\usepackage{threeparttable}

\title{From Handcrafted Features to Functional Edge Learning: Evolution of EEG Seizure Detection Frameworks}

\date{}

\author{Sepideh Kheirollahi \\
	School of Electrical and Computer Engineering \\
	College of Engineering, University of Tehran \\
	Tehran, Iran \\
	\texttt{s.kheirollahi@ut.ac.ir} \\
	\And
	Mohammad Rasoul Roshanshah \\
	School of Electrical and Computer Engineering \\
	College of Engineering, University of Tehran \\
	Tehran, Iran \\
	\texttt{mrroshanshah@ut.ac.ir} \\
}

\renewcommand{\shorttitle}{Evolution of EEG Seizure Detection Frameworks}

\hypersetup{
	pdftitle={Evolution of EEG Seizure Detection Frameworks}
}

\begin{document}
	\maketitle
	
	\begin{abstract}
		Electroencephalogram (EEG) analysis remains the clinical gold standard for epilepsy diagnosis and seizure detection. While Deep Learning (DL) has significantly advanced automated EEG interpretation, its transition from controlled experimental settings to routine clinical deployment is severely bottlenecked by fundamental architectural flaws. Standard DL models operate as opaque black-boxes lacking clinical interpretability, demand massive amounts of balanced annotated data, and incur steep computational costs incompatible with resource-constrained wearable or implantable neuromodulation devices. This paper presents a comprehensive review of these prevailing limitations and explores Kolmogorov-Arnold Networks (KANs) as a emerging paradigm for EEG-based seizure detection. By replacing the fixed activation functions of traditional neurons with flexible, learnable functions along the network’s connections, KANs bridge the critical gap between predictive accuracy and mathematical transparency. We systematically analyze how KAN architectures resolve the shortcomings of traditional DL-based models by offering exceptional parameter efficiency, inherent interpretability for physician trust, and robust performance under data scarcity. Ultimately, this review establishes KANs not merely as an incremental algorithmic update, but as a fundamental paradigm shift necessary to actualize next-generation, patient-specific, and thoroughly transparent clinical EEG monitoring systems.
	\end{abstract}
	
	\keywords{	
	\and Electroencephalography (EEG)
	\and Seizure Detection
	\and Kolmogorov-Arnold Networks (KANs)
	\and Explainable AI (xAI)
	\and Deep Learning
	}

	\section{Introduction}
	\label{sec:introduction}
	
	Epilepsy is a chronic neurological disorder affecting around 50 million people worldwide \cite{refnn1}. It is characterized by recurrent, unprovoked seizures, which can cause abnormal movements, sensory disturbances, or a loss of consciousness \cite{refnn1}. Early detection is critical; with a timely diagnosis and proper treatment, about 70\% of patients can achieve effective seizure control and significantly improve their quality of life \cite{refnn1}. Electroencephalography (EEG) is the clinical gold standard for monitoring this condition. It captures the distinct electrical patterns—such as spikes, sharp waves, and rhythmic discharges—that signal a seizure is occurring \cite{ref4}. However, traditional EEG analysis requires trained neurologists to manually review the recordings. This method is inherently time-consuming, vulnerable to inconsistent interpretations among clinicians, and highly impractical for long-term monitoring, which generates massive amounts of data \cite{refk-2}. Furthermore, because seizures can be rare, subtle, or happen during sleep, they are often difficult to identify through standard clinical observation alone. To overcome these challenges, automated seizure detection systems have been developed. By continuously analyzing EEG data and triggering real-time alerts, these systems enable faster medical intervention and greatly enhance patient safety \cite{ref99}.
	
	Over the past decade, advances in artificial intelligence have profoundly transformed these automated systems. Classical Machine Learning (ML) approaches—including Support Vector Machines (SVM) \cite{ref85}, K-Nearest Neighbors (KNNs) \cite{ref87}, and Decision Trees (DT) \cite{ref88}—established the foundation by relying on handcrafted features extracted from the time, frequency, or time-frequency domains. More recently, Deep Learning (DL) architectures such as Convolutional Neural Networks (CNNs) \cite{ref8}, Recurrent Neural Networks (RNNs) \cite{ref97}, and Transformer-based models \cite{ref98} have dominated the field. These advanced models perform hierarchical feature learning, enabling them to map the complex spatial and temporal dependencies inherent in non-stationary EEG recordings directly from raw or minimally processed data \cite{ref4}. 
	
	Despite reporting high classification accuracies in controlled settings, the routine clinical deployment of these standard DL models remains severely bottlenecked by fundamental structural limitations. First, massive DL architectures function as opaque black-boxes, lacking the interpretability required to earn physician trust, facilitate clinical debugging, or align with neurophysiological patterns \cite{ref23, ref52, ref61}. Second, they struggle with cross-patient generalization and robustness because they are highly sensitive to physiological artifacts, distribution shifts, and varying clinical hardware \cite{ref23, ref39, ref61, ref69}. Finally, relying on millions of static parameters demands massive, balanced datasets—which are incredibly scarce in epileptology—and incurs substantial computational costs that are fundamentally incompatible with resource-constrained environments like bedside monitors or implantable neuromodulation devices \cite{ref25, ref40, ref61}.
	
	Addressing these deep-rooted challenges necessitates a paradigm shift toward architectures that natively bridge the gap between theoretical expressiveness and practical clinical usability. Kolmogorov-Arnold Networks (KANs) represent a highly promising solution to this bottleneck \cite{refk-2, refk-13}. Moving away from the fixed activation functions of traditional models, KANs utilize learnable functions on network connections to create a transparent mathematical architecture, inherently boosting interpretability and facilitating the integration of domain knowledge \cite{refk-1, refk-4, refk-14}. Furthermore, KANs are parameter-efficient and robust to data scarcity, reducing the risk of overfitting on imbalanced EEG datasets \cite{refk-1, refk-2, refk-12}. Their flexible mathematical structure enables easier cross-patient personalization without exhaustive retraining, ensuring stable performance under distribution shifts and reliable deployment in real-world clinical and wearable environments \cite{refk-2, refk-11, refk-13}.
	
	This review provides a comprehensive, structured analysis of contemporary automated EEG-based seizure detection, with a specific focus on methodologies published within the past ten years. Unlike previous reviews that focus exclusively on conventional deep learning, we analyze traditional models alongside the KAN framework to assess their potential for deployment in clinical practice. The remainder of this paper is organized as follows: Section \ref{sec:background} outlines the medical and technical background of EEG. Section \ref{sec:pipeline} details the standard computational pipeline, followed by a review of classification models in Section \ref{sec:evolution}. Section \ref{sec:challenges} formalizes the critical barriers to clinical deployment. Section \ref{sec:kans} introduces the theoretical foundations and specific advantages of KANs. Finally, Section \ref{sec:future} proposes novel future research directions, and Section \ref{sec:conclusion} concludes the review.

	\section{Medical and Technical Background of EEG}
	\label{sec:background}
	
	The automated detection of epileptic seizures relies on a comprehensive understanding of both the physiological nature of brain activity and the technical parameters involved in its measurement. Before designing or evaluating computational models, it is essential to understand how EEG signals are generated, the environmental and biological noise that can corrupt them, and the standardized datasets used to train machine learning algorithms. Furthermore, establishing rigorous evaluation metrics is critical for comparing the efficacy of different detection frameworks. This section provides the foundational medical and technical concepts necessary for analyzing EEG-based seizure detection systems.
	
	\subsection{Principles of EEG and qEEG Analysis}
	EEG records voltage fluctuations generated by neuronal activity using scalp electrodes, capturing both the temporal and spatial dynamics of the brain. Abnormal patterns, such as those associated with epileptic seizures, often appear as distinctive rhythmic activity in the recordings \cite{ref6}. Each channel represents the voltage difference between electrodes, either relative to a reference electrode (reference montage) or between two active electrodes (bipolar montage). In most clinical applications and ML studies, the number of channels is typically considered equivalent to the number of electrodes when a common reference scheme is applied. In contrast, bipolar configurations pair electrodes together, producing a greater number of channels than electrodes alone \cite{ref16}. Different EEG recording protocols vary in electrode number and placement. The most widely used clinical method is the 10–20 system, in which electrodes are positioned at proportional distances across the scalp, providing 21 standard scalp recording sites \cite{ref54}. To achieve denser spatial sampling, the 10–10 system adds electrodes at intermediate positions, expanding the montage to 21–74 electrodes \cite{ref56, ref55}. Further extensions, such as the 10–5 system, support more than 300 electrode positions \cite{ref100}, enabling finer cortical coverage and supporting research and advanced computational approaches, including ML–based abnormality detection.
	
	According to Fig~\ref{fig:my_eeg_figure1} (a), EEG brain rhythms are commonly grouped into frequency bands associated with specific functional states. Delta activity (0.5–4 Hz) dominates during deep sleep, while theta waves (4–8 Hz) are linked to drowsiness and memory-related processes. The alpha band (8–12 Hz) is strongest during relaxed, non-demanding states, whereas beta activity (12–30 Hz) reflects active thinking and movement. Higher-frequency gamma oscillations (>30 Hz) are associated with complex processing and information integration across brain regions \cite{ref18}. The choice of sampling rate and frequency band in EEG analysis depends on the application. The sampling rate determines how accurately fast neural oscillations are captured and must be at least twice the highest frequency of interest to prevent distortion. Because most clinical and research EEG focuses on frequencies below 70–80 Hz, sampling rates of 200–256 Hz are usually sufficient, while higher rates of 500–1000 Hz are used when greater temporal accuracy is needed \cite{ref17}.
	
	qEEG applies computational analysis to recorded EEG signals to produce numerical measures and visual maps for objective interpretation \cite{ref11}. Compared with traditional EEG, qEEG adds frequency-band power analysis, signal complexity, connectivity, and network measures, enhancing consistency and revealing subtle abnormalities. These features make qEEG useful in disorders such as epilepsy, traumatic brain injury, Alzheimer’s disease, and certain psychiatric conditions. However, qEEG does not replace conventional EEG and is best used as a complementary tool \cite{ref10}.
	
	\subsection{Noise, Artifacts, and Signal Degradation}
\begin{figure}[t]  
		\centering
		\includegraphics[width=0.7\textwidth]{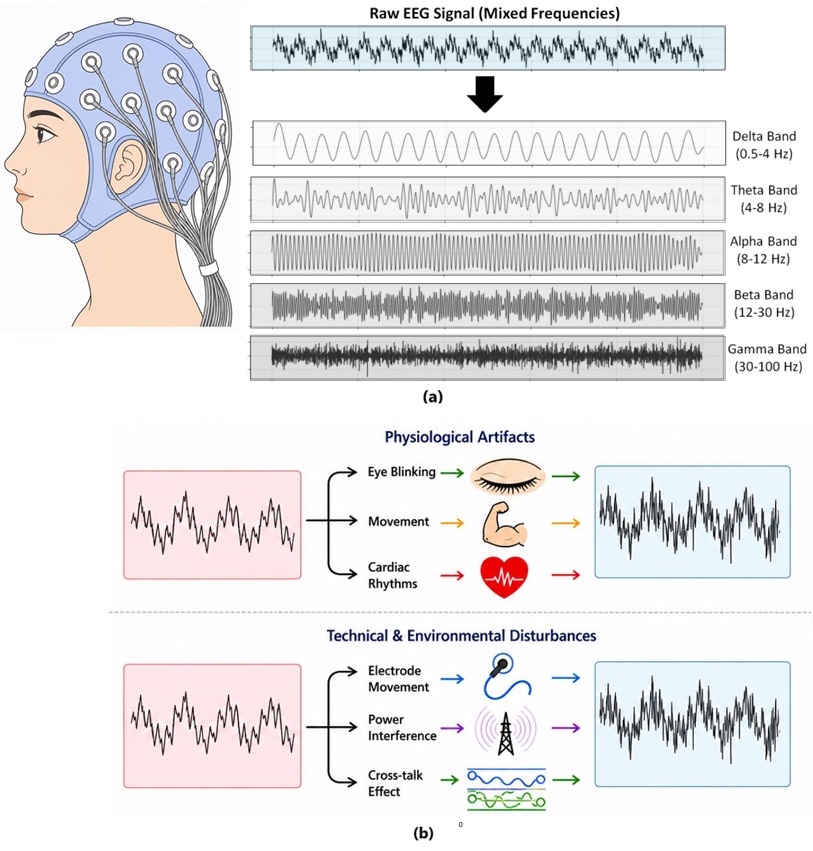}
		\caption{(a) Illustration of scalp EEG acquisition and its main frequency bands: delta, theta, alpha, beta, and gamma. (b) Overview of physiological and technical artifacts that distort EEG signals, including eye blinks, movement, cardiac activity, electrode shifts, power‑line noise, and cross‑talk.}
		\label{fig:my_eeg_figure1}
	\end{figure}

	Although EEG is a valuable tool for monitoring brain activity, its low amplitude makes it highly prone to noise \cite{ref19}. Such disturbances can mask relevant signals and affect clinical or research conclusions. As shown in Fig~\ref{fig:my_eeg_figure1} (b), noise sources fall into two main categories: physiological artifacts and technical or environmental disturbances. Physiological artifacts come from bodily functions that produce stronger electrical signals than the brain, such as eye blinks, movements, muscle contractions, and cardiac rhythms. Technical and environmental artifacts arise from external factors, including mains power interference, electrode instability or movement, and cross-talk effect in cables or recording equipment \cite{ref20}.
	
	\subsection{Standardized Datasets in Seizure Detection}
	Seizure detection research relies heavily on EEG datasets, which can be broadly divided into publicly available research datasets and clinical datasets collected in healthcare settings. The choice between these two categories reflects a trade-off between accessibility and standardization versus the ability to capture realistic clinical conditions \cite{ref60}.
	
	EEG datasets can be classified according to several criteria, one of the most important being the signal acquisition
	method, which distinguishes non-invasive scalp EEG recordings from invasive intracranial EEG (iEEG) recordings obtained using surgically implanted electrodes  \cite{ref61}. Compared with scalp EEG, iEEG typically offers a higher signal-to-noise ratio and fewer artifacts. Datasets are also categorized by temporal characteristics, including continuous recordings—either long-term (exceeding 24 hours) or short-term—and non-continuous datasets composed of randomly segmented EEG samples. In addition, electrode placement schemes, most often following the international 10–20 system, are critical for ensuring repeatability and standardized feature extraction and analysis \cite{ref60}.
	
	Public repositories such as the CHB-MIT, Bonn, and TUH EEG Corpus (TUSZ) are widely used because they are accessible, well documented, and provide standardized benchmarks for algorithm development \cite{ref39}. The CHB-MIT database is frequently employed for long-term, continuous scalp EEG analysis, with structured recordings that support the evaluation of extended monitoring algorithms. In contrast, the Bonn dataset contains short, carefully curated EEG segments. Its small size and the absence of patient‑specific identifiers limit how well models trained on it can generalize to real clinical settings. However, because the signals are clean and well‑annotated, the dataset is highly suitable for early-stage feature extraction and proof‑of‑concept studies. TUSZ distinguishes itself through its large scale and longitudinal design, encompassing recordings from more than 600 subjects, and is therefore essential for training robust DL models capable of capturing complex, real-world EEG patterns while remaining more standardized than most private clinical collections \cite{ref39, ref60,ref23}. 
	
	By contrast, clinical datasets are acquired directly from hospital or diagnostic environments and exhibit greater variability in recording conditions, patient populations, and seizure types, making them more representative of real-world applications \cite{ref62}. Notable examples include long-term recordings from epilepsy monitoring units and trial datasets such as the NeuroVista \cite{ref63} study, which provide high-resolution, long-term iEEG data. These datasets offer the signal quality and data volume required for advanced algorithm validation, particularly for seizure detection systems intended for use outside controlled hospital settings \cite{ref40, ref41}.
	In Table~\ref{tab:eeg_comparison}, the features of these datasets are briefly classified.

	\begin{table}[!htbp]
	    \caption{Comparison of Key Features of EEG Datasets for Seizure Detection.}
	    \label{tab:eeg_comparison}
	    \footnotesize
	    \centering
	    
	    \begin{tabular}{
	            p{2.0cm} 
	            p{2.0cm} 
	            p{2.0cm} 
	            p{2.5cm} 
	            p{1.2cm} 
	            p{1.5cm} 
	        }
	        \toprule
	        \textbf{Dataset} & 
	        \textbf{Seizure Types} & 
	        \textbf{Duration} &
	        \textbf{Channels/\newline Sampling Rate} & 
	        \textbf{Artifact \& Noise} & 
	        \textbf{Data\newline Accessibility}  \\
	        \midrule
	        
	        CHB-MIT \cite{refD1} & 
	        \makecell[l]{\text{Generalized} \\ \text{Focal}} & \makecell[l]{Long-term \\($\sim$1h/Segment)} & 18-24 Channels 256 Hz &  Not Clean & {\hspace{0pt}}Public \\ \\
	        
	        Bonn \cite{refD2} & 
	        Healthy, Interictal, Ictal & \makecell[l]{Short-term\\ ($\sim$23.6s/ Segment)} & \makecell[l]{Single Channel \\ 173.61 Hz} &  Clean & {\hspace{0pt}}Public \\ \\
	        
	        UCI \cite{refD7} &
	        \makecell[l]{Healthy, Interictal,\\ Ictal} & \makecell[l]{Short-term\\ ($\sim$23.6s/ Segment)} & 
	        \makecell[l]{Single Channel \\ 173.61 Hz}
	        & Clean & \makecell[l]{Public} \\ \\
	        
	        NICU(Helsinki)\newline \cite{refD6} &
	        \makecell[l]{Ictal, \\ Interictal} &
	        \makecell[l]{Short-term, \\ (1-2h/ Segment)} & 
	        \makecell[l]{19 Channels \\ 256 Hz}
	        & Not Clean & \makecell[l]{Public} \\ \\
	        
	        TUH \cite{refD3} &
	        \makecell[l]{Generalized, \\ Focal, Complex} & 
	        \makecell[l]{Long-term, \\ (< 1h/ Segment)}
	        & \makecell[l]{23–31 Channels\\ 250–400 Hz} & Not Clean &  \makecell[l]{Public} \\ \\
	        
	        HCTM  \cite{ref71} &
	        \makecell[l]{Focal} & \makecell[l]{Long-term, \\ Varied} & 
	        \makecell[l]{16 Channels \\ 500 Hz}
	        & Not Clean & \makecell[l]{Controlled-\\access} \\ \\
	        
	        EPILEPSIAE\newline \cite{ref41} & 
	        Focal & \makecell[l]{Long-term, \\ Varied} & 
	        \makecell[l]{Up to 128 Channels \\ 256–2500 Hz} & Not Clean &
	        \makecell[l]{Controlled-\\access} \\ \\
	        
	        SWEC-ETHZ\newline \cite{refD5} & 
	        Focal & \makecell[l]{Long-term \\($\sim$1h/ Segment)} & 24–128 Channels 512 or 1024 Hz & Mostly clean with minor noise & \makecell[l]{Controlled-\\access} \\
	        
	        \bottomrule
	    \end{tabular}
	\end{table}
	
	\subsection{Performance Evaluation Metrics}
	Assessing the effectiveness of ML and DL models in accurately detecting seizures is crucial for reliable EEG-based diagnosis. To evaluate these models, several performance criteria are commonly used, which are summarized in the Table~\ref{tab:evaluation}. These metrics are based on four possible outcomes of the model’s predictions: True Positive (TP), True Negative (TN), False Positive (FP), and False Negative (FN). TP represents cases in which the model correctly identifies a seizure. TN corresponds to correctly recognized non-seizure events. FP occurs when the model incorrectly classifies a non-seizure event as a seizure, while FN represents missed seizures that are incorrectly labeled as non-seizure events. These fundamental categories form the basis for calculating key evaluation metrics such as accuracy, sensitivity, specificity, and precision, providing a comprehensive view of the model’s diagnostic performance \cite{ref33, ref34}. In addition, to evaluate how strongly a model’s performance depends on individual patient data, various validation strategies are employed. These include statistical techniques such as $k$-fold cross-validation, Leave-One-Subject-Out (LOO) validation, and strict separation between training and testing datasets to ensure that evaluation is performed on unseen patient recordings \cite{ref35}.
	
	\begin{table}[!htbp]
	    \caption{Evaluation Metrics for EEG-Based Seizure Detection.}
	    \label{tab:evaluation}
	    \footnotesize
	    \renewcommand{\arraystretch}{1.5} 
	    \centering
	    
	    \begin{tabular}{
	            p{2.5cm} 
	            p{5.0cm} 
	            p{4.5cm} 
	        }
	        \toprule
	        \textbf{Metric} & \textbf{Formula} & \textbf{Description} \\
	        \midrule
	        
	        Accuracy & \small
	        $\frac{TP + TN}{TP + TN + FP + FN}$ & 
	        Overall proportion of correct predictions. \\ \\
	        
	        Sensitivity (Recall) & 
	        $\frac{TP}{TP + FN}$ & 
	        Fraction of actual seizures correctly detected. \\ \\
	        
	        Specificity & 
	        $\frac{TN}{TN + FP}$ & 
	        Fraction of non-seizure frames correctly detected. \\ \\
	        
	        Precision & 
	        $\frac{TP}{TP + FP}$ & 
	        Fraction of predicted seizures that are correct. \\ \\
	        
	        F1 Score & 
	        $2 \cdot \frac{Precision \cdot Sensitivity}{Precision + Sensitivity}$ & 
	        Balance between precision and recall; useful for imbalanced datasets. \\ \\
	        
	        \makecell[l]{False Positive Rate\\ (FPR)} & 
	        $\frac{FP}{FP + TN}$ & 
	        Proportion of non-seizure events incorrectly classified as seizures; indicates the likelihood of false alarms. \\ \\
	        
	        \makecell[l]{True Positive Rate \\(TPR)} & 
	        $\frac{TP}{TP + FN}$ & 
	        Proportion of actual seizure events correctly classified as seizures; indicates the model's ability to detect real seizures. \\ \\
	        
	        \makecell[l]{False Alarm Rate\\ (FAR)} & 
	        $\frac{FP}{\text{Total duration or total non-seizure events}}$ & 
	        Rate of false seizure detections per time unit or event. \\ \\
	        
	        \makecell[l]{ROC-AUC} &
	        \makecell[l]{\small
	        	$\displaystyle \int_{0}^{1}\mathrm{TPR}(\mathrm{FPR})\,d(\mathrm{FPR})$
	        } &
	        Threshold-independent measure of the model's ability to discriminate between seizure and non-seizure classes. (more common) \\ \\
	        
	        \makecell[l]{PR-AUC} &
	        \makecell[l]{\small
	        	$\displaystyle
	        	\int_{0}^{1}\mathrm{Precision}(\mathrm{Recall})\,d(\mathrm{Recall})$
	        } &
	        Measures the trade-off between precision and recall across all decision thresholds. \\

	        \bottomrule
	    \end{tabular}
	\end{table}

	\section{The Standard Computational Pipeline for Seizure Detection}
	\label{sec:pipeline}
	
\begin{figure}[t]  
		\centering
		\includegraphics[width=0.7\textwidth]{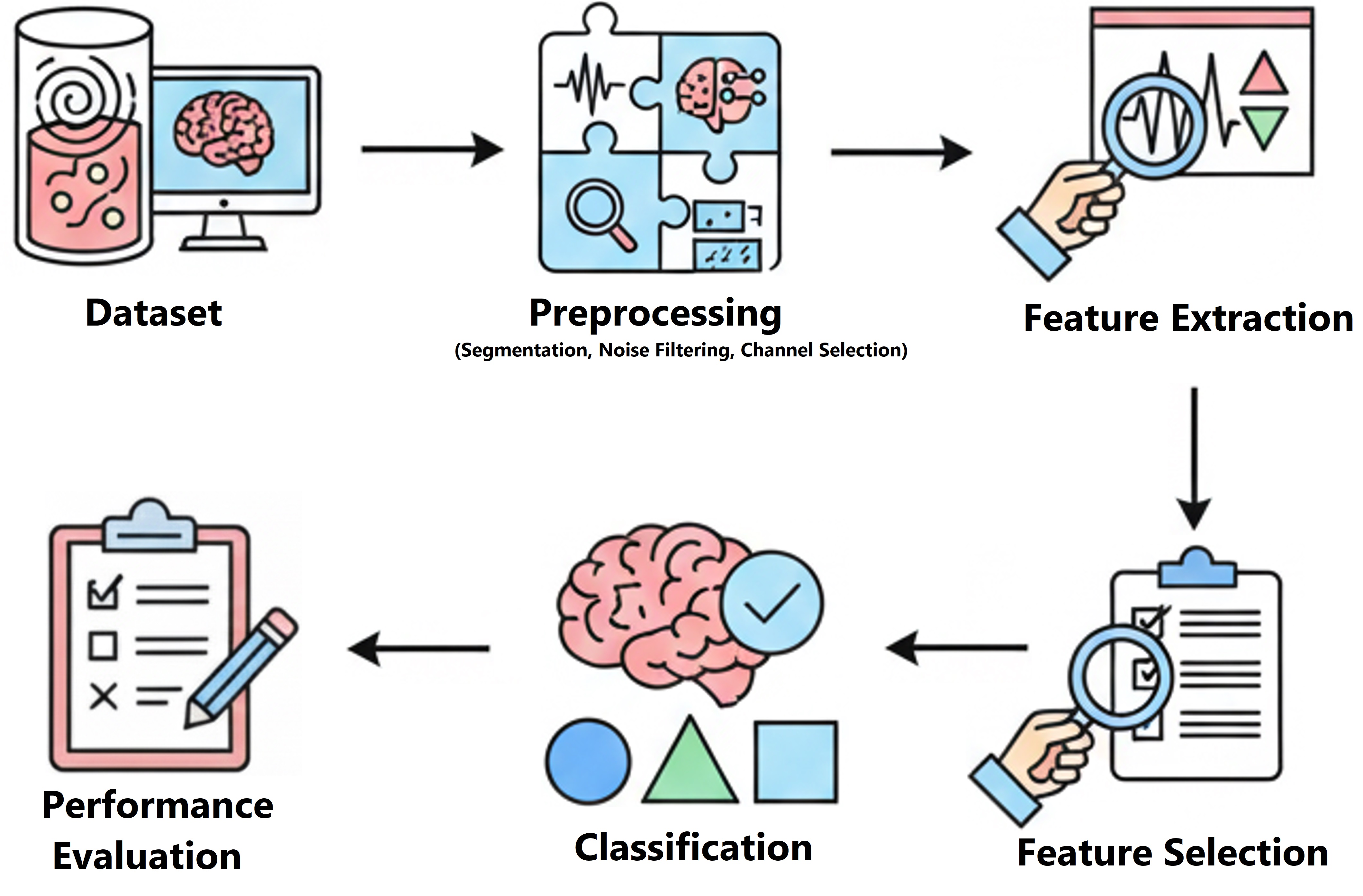}
		\caption{Overall pipeline for data classification, progressing from dataset preparation and preprocessing to feature extraction, feature selection, classification, and final performance evaluation.}
		\label{fig:my_eeg_figure2}
	\end{figure}

	Raw EEG signals are inherently complex, highly dimensional, and heavily contaminated by noise, making them generally unsuitable for direct ingestion by classification algorithms. Consequently, automated seizure detection frameworks typically follow a structured computational pipeline designed to distill raw continuous data into actionable diagnostic decisions. According to Fig \ref{fig:my_eeg_figure2},this pipeline generally consists of three sequential stages: data preprocessing to enhance the signal-to-noise ratio, feature extraction to map the signals into discriminative mathematical representations across various domains, and dimensionality reduction to optimize computational efficiency. This section outlines the standard methodologies employed at each stage of this pipeline.
	
	\subsection{Signal Preprocessing and Artifact Removal}
	EEG recordings are often contaminated by artifacts and external noise, making raw signals difficult to analyze directly \cite{ref24}. EEG characteristics also vary across subjects and electrode channels. Seizure events occur far less frequently than non-seizure activity, causing severe class imbalance during model training \cite{ref25}. For these reasons, preprocessing is essential. It reduces noise and normalizes inter-subject and inter-channel variability by placing signals on a comparable scale. These steps improve input consistency, enhance feature quality, and increase the accuracy of seizure detection models \cite{refk-26}. This phase involves a variety of techniques, including data cleaning, standardization, resampling, segmentation, and dimensionality reduction \cite{ref94}. The cleaning stage focuses on removing noise and artifacts introduced during the recording of the original signal. Depending on the analysis domain, which may be time, frequency, or time-frequency, different methods can be applied, such as Butterworth filtering \cite{refn7}, FIR filtering, Independent Component Analysis (ICA) \cite{ref23}, Local Mean Decomposition (LMD) \cite{ref79}, the fourier transform, Short-Time Fourier Transform (STFT), the wavelet transform \cite{refn8} and the S‑transform \cite{ref75}. Additional techniques, including simple moving average, exponential moving average  \cite{refn9}, artifact subspace reconstruction, and channel-reduced differential \cite{refk-26}, also contribute to smoothing and enhancing the overall quality of the data. 
	
	Since the values of different variables can cover a wide range, determining an optimal point becomes challenging. Therefore, normalization of both the training and testing datasets is essential. Common normalization approaches include Z‑scores and Min‑Max scaling. Furthermore, when datasets are recorded with different sampling rates, inconsistencies may arise that negatively affect analytical accuracy. In such cases, resampling techniques are applied during the pre‑processing stage to unify the sampling rates across datasets \cite{ref76}.
	
	EEG signals are inherently non‑stationary, meaning that their statistical characteristics change over time \cite{ref81}. To address this issue, segmentation is often applied during the pre‑processing stage, where long EEG recordings are divided into shorter intervals that exhibit quasi‑stationary behavior and maintain more consistent statistical properties in both the time and frequency domains \cite{refn3}. These intervals may be either overlapping or non‑overlapping.
	
	In addition, an imbalance between epileptic and non‑epileptic samples can hinder effective model training. To overcome this issue, oversampling techniques such as  Synthetic Minority Oversampling Technique (SMOTE) \cite{refn4} and  K-nearest
	neighbor sampling approach (KNNOR) \cite{refn5} may be applied during the pre‑processing stage to increase the number of epileptic instances \cite{ref69}. Finally, dimensionality‑reduction algorithms such as Principal Component Analysis (PCA) can be used to reduce data complexity by removing redundant information, which in turn decreases computational cost and improves the efficiency of subsequent analysis steps.
	
	\subsection{Feature Extraction Domains (Time, Frequency, and Time-Frequency)}
	An essential step in EEG-based epilepsy diagnosis is feature extraction from raw recordings that can be the time domain, frequency domain, or both \cite{ref3}. Appropriate features directly affect the accuracy and robustness of automated detection systems. Conventional methods rely on handcrafted features, such as simple statistical descriptors (e.g., mean and variance) that summarize amplitude distributions \cite{ref12}. Nonlinear measures are also used to capture the complex and irregular dynamics of brain activity. Signal decomposition techniques, including the wavelet transform and empirical mode decomposition, reorganize EEG signals into time–frequency components. This process reveals transient events and frequency-specific patterns that are often hidden in the raw signal \cite{ref13}.
	
	In many practical applications, decision-making relies on extracting informative features from raw signals. Feature extraction methods are commonly categorized into time-domain, frequency-domain, and time–frequency approaches. Time-domain features are derived directly from amplitude variations over time, whereas frequency-domain features are obtained by transforming the signal, for example using the Fourier transform, to describe energy distribution across frequency bands. Time–frequency approaches combine both perspectives to capture how spectral content evolves over time \cite{ref57}.
	
	The choice among these domains depends on signal stationarity and available computational resources \cite{ref58}. Stationary signals exhibit time-invariant statistical properties, such as mean, variance, and frequency content, making frequency-domain techniques—particularly power spectral density analysis—well suited for revealing their spectral structure \cite{ref22, ref59}. In contrast, non-stationary signals have time-varying statistical and spectral characteristics, for which conventional Fourier methods are limited due to their global frequency representation. Time–frequency approaches, such as the STFT and wavelet transform, are therefore more appropriate, as they capture the temporal evolution of frequency content \cite{ref21}.
	
	In contrast, DL architectures automatically learn informative representations directly from EEG data. Convolutional and recurrent hidden layers extract hierarchical features without explicit manual design. This reduces dependence on handcrafted feature extraction and selection. As a result, DL enables a unified and automated processing pipeline. In most cases, only minimal preprocessing is required, such as band-pass filtering or time–frequency representations like the discrete wavelet transform (DWT) and STFT. These properties make DL approaches well suited for real-time seizure detection \cite{ref1}.
	
	\subsection{Feature Selection and Dimensionality Reduction}
	Because feature extraction in traditional ML often produces a large number of candidates, an additional feature selection or dimensionality reduction stage is required. Feature selection identifies the most relevant subset for distinguishing seizure from non-seizure states, reducing computational cost and the risk of over-fitting. Common approaches include filter-based, wrapper-based, and embedded methods \cite{ref27}. Alternatively, dimensionality reduction techniques such as PCA \cite{ref14}, Linear Discriminant Analysis (LDA) \cite{ref15}, uniform manifold approximation and projection \cite{ref84}, auto-encoders \cite{ref83}, and t-distributed stochastic neighbor embedding \cite{ref96} compress the feature space while preserving essential information \cite{ref28, ref29}. In contrast, DL models automatically learn relevant representations directly from EEG data. Feature extraction and selection are performed implicitly by the network, removing the need for handcrafted features and separate selection stages \cite{ref1}.
	
	\subsection{Classification Methodologies}
	\label{sec:classification}
	
	Historically, the final diagnostic stage relied on classical ML models to categorize EEG signals into seizure or non-seizure states. These approaches range from foundational linear models like logistic regression \cite{ref86} to robust ensemble classifiers such as Random Forests (RF) \cite{ref89}, gradient boosting \cite{ref90}, XGBoost \cite{ref91}, and Light Gradient Boosting Decision Tree (LightGBM) \cite{ref92}. While these traditional algorithms are effective in controlled scenarios, their fundamental limitation is an absolute dependence on the subjective and computationally expensive process of handcrafted feature extraction.
	
\begin{figure}[t]
		\centering
		\includegraphics[width=\textwidth]{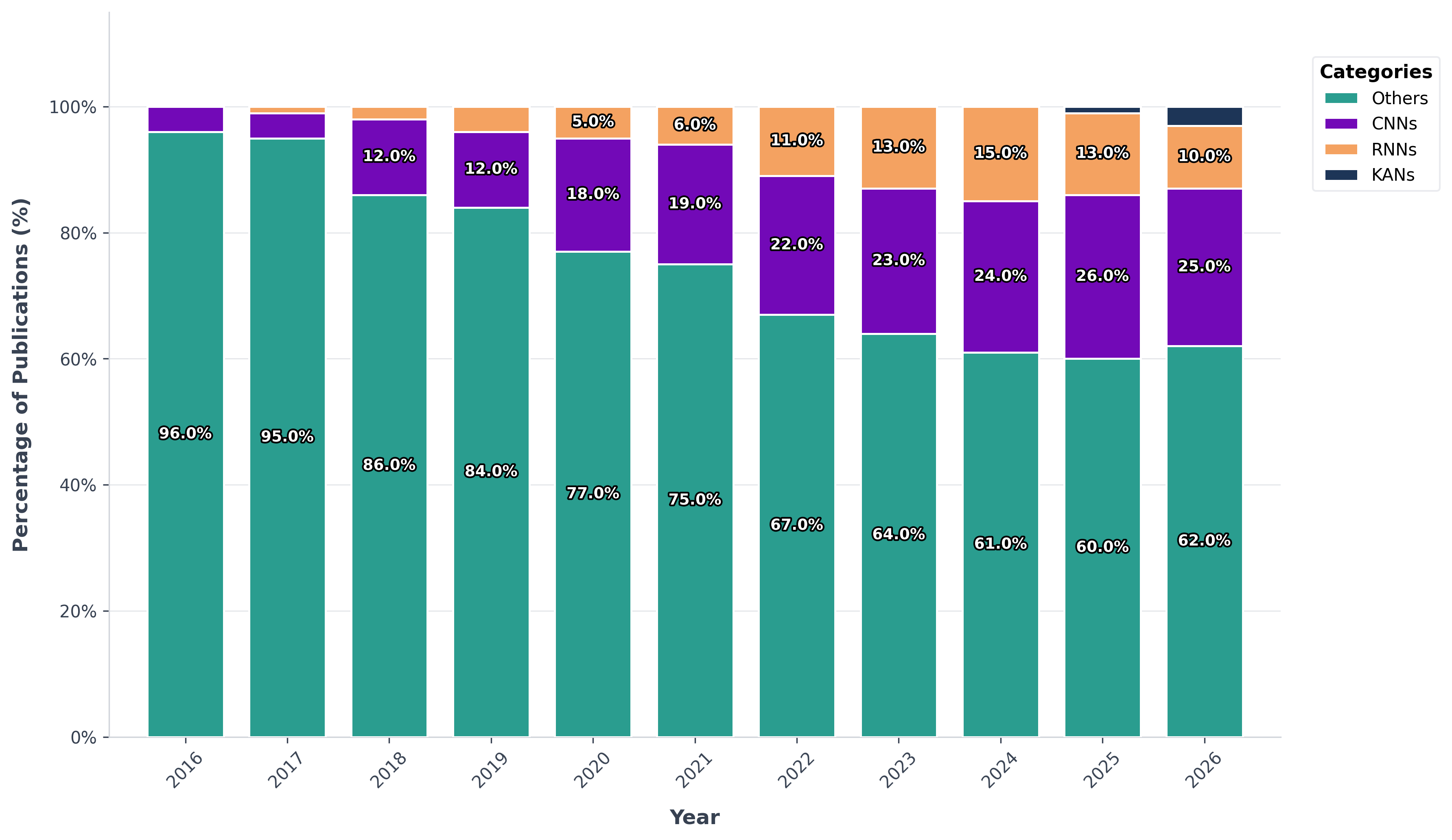}
		\caption{
			Relative proportions of published research methodologies for EEG-based seizure detection (2016--2026). 
			Data extracted from the Scopus database (May 2026).
		}
		\label{fig:publication_trends}
	\end{figure}

	To avoid the difficult task of creating features manually, modern research has shifted toward deep learning. These models simplify the process by automatically finding patterns and classifying the signals simultaneously. As illustrated in Figure \ref{fig:publication_trends}, bibliometric data extracted from Scopus demonstrates a clear historical shift in classification methodologies, highlighted by a rapid and sustained proliferation of CNN and RNN research since 2018. Within this domain, CNNs are predominantly utilized to capture spatial relationships across multiple EEG electrodes, whereas RNNs, particularly Long Short-Term Memory (LSTM) \cite{ref93} networks, are specialized for modeling the long-term sequential dependencies inherent in non-stationary brain activity. By merging these capabilities, hybrid models such as CNN-LSTM \cite{ref94} and Bidirectional LSTM (BiLSTM) \cite{ref95} provide a comprehensive framework capable of mapping complex spatio-temporal correlations directly from raw data. 
	
	However, while these standard DL models achieve remarkable diagnostic accuracy, their massive parameter counts and opaque, black-box decision-making processes create significant clinical bottlenecks. These unresolved challenges have ultimately catalyzed the very recent emergence of transparent, parameter-efficient alternatives like KANs, which, as evidenced by the 2026 publication data, are now beginning to establish a vital new paradigm in the seizure detection landscape.
	
	\section{Evolution of Classification Models in Seizure Detection}
	\label{sec:evolution}
	
	The classification stage serves as the core decision-making engine of any automated seizure detection framework. Over the past decade, the landscape of classification algorithms has undergone a significant paradigm shift. Early systems relied primarily on traditional ML models, which required meticulous, handcrafted feature engineering. More recently, DL architectures have dominated the field, offering the ability to automatically learn complex spatio-temporal representations directly from minimally processed data. This section provides a comprehensive review of the existing literature, categorizing state-of-the-art methodologies into traditional ML frameworks and DL approaches, including Spatial, Temporal, and Hybrid architectures.
	
	The selection between traditional ML models and DL architectures is largely influenced by factors such as dataset size, the complexity of extracted features, and computational limitations (real-time or offline analysis). For small- to medium-scale datasets, conventional classifiers are often more practical, as they balance accuracy with efficiency \cite{ref64}. In contrast, DL models are typically favored when working with large datasets, where their ability to learn complex patterns outweighs their higher computational demands \cite{ref65}.
	
	\subsection{Traditional Machine Learning Approaches}
	In traditional ML for EEG-based seizure detection, extracted features are analyzed to determine whether a signal corresponds to a seizure or non-seizure state. Existing approaches are generally grouped according to their modeling principles, including statistical and linear models, distance- and instance-based methods, and tree-based or ensemble classifiers.  
	
	\begin{itemize}
		\item \textbf{Statistical and Linear Models:} These methods model direct, often linear, relationships between EEG features and seizure classes. Techniques such as SVM \cite{ref85}, logistic regression \cite{ref86}, and LDA \cite{ref15} are widely used due to their simplicity and low computational cost. They perform well when features clearly separate interictal and ictal states, making them suitable for real-time or hardware-oriented detection systems.
		
		\item \textbf{Distance- and Instance-Based Methods:} Unlike statistical models, these methods classify samples based on similarity in the feature space. The KNN algorithm assigns a class label according to the majority among the $k$ closest labeled samples, using distance metrics such as Euclidean, Manhattan, or Mahalanobis distance \cite{ref87}. This data-driven approach can handle complex and nonlinear feature distributions without explicit training, though performance depends on the choice of metric and $k$, and inference can be memory- and computation-intensive.
		
		\item \textbf{Tree-Based and Ensemble Methods:} These approaches capture nonlinear relationships by dividing the feature space into hierarchical, rule-based regions. DT form the basic model but can overfit. Ensemble methods, including RF \cite{ref89} and gradient boosting \cite{ref90}, combine multiple learners to improve robustness and stability. Advanced implementations such as XGBoost \cite{ref91} and LightGBM \cite{ref92} enhance accuracy and efficiency. These methods are flexible, handle noisy or variable EEG data well, and are suitable for real-time or hardware-based seizure detection, though they can be computationally demanding.
	\end{itemize}
	In the following we reviewed several papers that used traditional ML models for seizure detection. In addition all these reviewed papers are inserted briefly in Table~\ref{tab:ml_eeg_summary}.
	
	\begin{table}[!htbp]
        \centering 
        
        \begin{threeparttable} 
            \caption{Summary of Traditional ML-Based EEG Seizure Detection Studies.}
            \label{tab:ml_eeg_summary}
            \footnotesize
            \renewcommand{\arraystretch}{1.2}
            \setlength{\tabcolsep}{4pt}
            
            \begin{tabular}{
                    p{1.8cm} 
                    p{1.0cm} 
                    p{4.3cm} 
                    p{1.2cm} 
                    p{3.0cm} 
                }
                \toprule
                \textbf{Study} & \textbf{Dataset} & \makecell[l]{\textbf{Preprocessing /}\\ \textbf{Feature Extraction (Selection)}} & \textbf{Classifier} & \textbf{Key Results (\%)} \\
                \midrule
                
                Tran et al.\newline 2022\,\cite{ref7} &
                Bonn &
                \makecell[l]{Band-pass filtering(0.53–40\,Hz);\\
                    Label adjustment;\\
                    DWT (db4) and Sub-band selection;\\
                    statistical features; BPSO} &
                SVM, KNN, DT, RF &
                \makecell[l]{ACC - SEN - SPE \\
                    98.40 - 96.00 - 99.00 \tnote{$\dagger$} \\
                    97.60 - 92.00 - 99.00 \tnote{$\ddagger$} \\
                    96.00 - 96.00 - 97.00 \tnote{$\mathsection$}  \\
                    60.80 - 96.00 - 97.00 \tnote{$\mathparagraph$} } \\ \\
                
                Mahjoub et al.\newline 2019\,\cite{ref43} &
                Bonn &
                Window segmentation;\newline Signal decomposition\newline(TQWT + MEMD) &
                SVM &
                \makecell[l]{ACC - SEN - SPE \\ 
                    98.78 - 96.56 - 99.33 \tnote{$\dagger$} }\\ \\
                
                Mardini et al.\newline 2020\,\cite{ref44} &
                Bonn &
                \makecell[l]{Band-pass filtering; \\
                MAV, AVP, SD, Variance, Mean, \\Skewness
                Shannon Entropy, \\Max/Max, Normalized SD, Energy} &
                \makecell[l]{SVM,\\ ANN,\\ KNN} &
                \makecell[l]{ACC - SEN - SPE \\ 
                    98.00 - 97.50 - 100 \tnote{$\dagger$} \\
                    98.60 - 98.30 - 100 \tnote{$\parallel$} \\
                    98.60 - 98.30 - 100 \tnote{$\ddagger$} } \\ \\
                
                Kavitha et al.\newline 2022\,\cite{ref45} &
                Bonn &
                \makecell[l]{DWT (6 levels decomposition); \\
                Max/Min Coefficient, MAV, \\Std Dev, Avg Power,
                \\Approx. Entropy, Shannon Entropy} &
                KNN, SVM, DT &
                \makecell[l]{ACC - SEN - SPE \\ 
                    97.00 - 93.00 - 98.00 \tnote{$\ddagger$, $*$} \\
                    97.40 - 97.00 - 97.50 \tnote{$\dagger$, $*$} \\
                    79.75 - 74.50 - 85.00 \tnote{$\mathsection$, $*$} } \\ \\
                
                Shen et al.\newline 2022\,\cite{ref46} &
                CHB-MIT &
                \makecell[l]{30-s window segmentation; \\
                    DWT (6 levels decomposition);\\
                    SD, mean, max, median, kurtosis, \\band power
                    SE, LE, FE} &
                SVM &
                \makecell[l]{ACC - SEN \\
                    97.00 - 96.67 \tnote{$\dagger$} } \\ \\
                
                Zubair et al.\newline 2021\,\cite{ref47} &
                Bonn &
                \makecell[l]{DWT (5 levels decomposition);\\
                Energy, SR, RSA (time-domain)\\
                DF, SE, (frequency domain)\\
                SPPCA; SUBXPCA} &
                SVM, KNN, RF, MLP &
                \makecell[l]{ACC - PRE - F1 \\
                    96.6 - 100 \hspace{2pt}- 96.6 \tnote{$\dagger$, $\star$} \\
                    83.3 - 100 \hspace{2pt}- 85.2 \tnote{$\ddagger$, $\star$} \\
                    96.6 - 100 \hspace{2pt}- 96.6 \tnote{$\mathparagraph$, $\star$}  \\
                    76.6 - 75.8 - 80.0 \tnote{$\sharp$, $\star$} \\
                    93.3 - 100 \hspace{2pt}- 93.5 \tnote{$\dagger$, $\diamond$} \\
                    83.3 - 100 \hspace{2pt}- 89.2 \tnote{$\ddagger$, $\diamond$} \\
                    98.3 - 100 \hspace{2pt}- 98.3 \tnote{$\mathparagraph$, $\diamond$}  \\
                    91.6 - 93.1 - 92.0 \tnote{$\sharp$, $\diamond$} } \\ \\
                
                Savadkoohi et al.\newline 2020\,\cite{ref82} &
                Bonn &
                \makecell[l]{TD (Butterworth); FD (FFT);\\ TFD (DWT); T-test; SFFS} &
                SVM, KNN &
                \makecell[l]{ACC - SEN - SPE \\ 
                    98.0 - 98.7 - 95.0 \tnote{$\dagger$, $\triangle$} \\
                    97.2 - 98.5 - 92.2 \tnote{$\ddagger$, $\triangle$} \\
                    100 \hspace{2pt}- 100 \hspace{2pt}- 100 \tnote{$\dagger$, $\nabla$} \\
                    97.0 - 98.0 - 93.0 \tnote{$\ddagger$, $\nabla$} \\
                    99.8 - 100 \hspace{2pt}- 99.0 \tnote{$\dagger$, $\infty$} \\
                    97.2 - 98.2 - 93.0 \tnote{$\ddagger$, $\infty$} } \\
                
                \bottomrule
            \end{tabular}
            
            \begin{tablenotes}\footnotesize
                \item[$*$] Results correspond to the ABCD\_E dataset configuration.
                \item[$\dagger$] Results obtained using the SVM classifier.
                \item[$\ddagger$] Results obtained using the KNN classifier.
                \item[$\mathsection$] Results obtained using the DT classifier.
                \item[$\mathparagraph$] Results obtained using the RF classifier.
                \item[$\parallel$] Results obtained using the ANN classifier.
                \item[$\sharp$] Results obtained using the MLP classifier.
                \item[$\star$] Results obtained using SPPCA for feature extraction.
                \item[$\diamond$] Results obtained using SUBXPCA for feature extraction.
                \item[$\triangle$] Results obtained using TD (Butterworth) features.
                \item[$\nabla$] Results obtained using FD (FFT) features.
                \item[$\infty$] Results obtained using TFD (DWT) features.
            \end{tablenotes}
        \end{threeparttable} 
    \end{table}

	Tran et al.~\cite{ref7} developed an ML‑based platform that extracts features from raw EEG signals using the DWT transform. The binary particle swarm optimization algorithm then selects an optimal feature subset, reducing dimensionality without losing critical information. The extracted features are used to train models for classifying EEG signals into seizure and non-seizure categories. Among the models, SVM is employed, with performance further improved through hyper-parameter optimization.
	
	 Mahjoub et al.~\cite{ref43} proposed a unified framework that combines advanced EEG analysis with classical machine learning to build an adaptive system for seizure detection. To handle the nonlinear and non-stationary nature of epileptic EEG signals, the study uses the Tunable Q-Wavelet Transform (TQWT) to decompose signals into subbands with adjustable Q-factors, offering greater precision than fixed-parameter wavelets. Multivariate Empirical Mode Decomposition (MEMD) is also applied to multichannel EEG to extract intrinsic mode functions, preserving spatial relationships across brain regions. The framework provides a customizable decomposition pipeline, allowing selection of direct analysis, TQWT, or MEMD based on clinical needs. TQWT is suited for high-frequency resolution, while MEMD preserves spatial correlations. Features extracted from these methods are then classified using an SVM, which effectively handles the high-dimensional feature space and distinguishes seizure from non-seizure activity.
	
	Mardini et al.~\cite{ref44} proposed frameworks that aim to achieve both high detection accuracy and computational efficiency in EEG‑based seizure recognition. The pipeline starts with band-pass filtering to remove noise and artifacts, followed by DWT decomposition into multiple frequency bands to isolate seizure-related activity, often in higher frequencies. Fifty-four mother wavelets from different families are systematically evaluated to select the best representation. Handcrafted statistical features, such as absolute mean and variance, are extracted from the subbands, and a genetic algorithm identifies the most informative subset for seizure discrimination. These features are then classified using standard ML models, including SVM, KNN, and ANN. 
	
	Kavitha et al.~\cite{ref45} proposed a wavelet-based machine learning approach for automatic seizure detection. The DWT decomposes EEG signals into six frequency subbands, from which twelve time- and frequency-domain features such as mean absolute value, maximum and minimum coefficients, and standard deviation are initially computed. To reduce computational cost and improve generalization, only the seven most informative features are retained. These features are then used with classifiers including SVM, KNN, naïve bayes, and DT to distinguish seizure from non-seizure EEG segments.
	
	 Shen et al.~\cite{ref46} proposed a real-time seizure detection framework targeting high accuracy and computational efficiency for long‑term EEG monitoring. EEG signals are first decomposed into multiple sub-frequency bands using DWT, and eigenvalue-based features from eight algorithms are concatenated to form the final feature vector. The framework is evaluated in two phases. First, the discriminative power of the features is tested on a short-term UB dataset using an SVM to classify healthy, seizure-free, and seizure-active segments. For real-time evaluation, the CHB-MIT dataset is used on continuous long-term recordings. To address the imbalance between seizure and non-seizure windows, an ensemble of DTs with random undersampling of the majority class is employed, improving seizure sensitivity while maintaining a reasonable false-alarm rate.
	
	 Zubair et al.~\cite{ref47} proposed an machine learning pipeline that combines time and frequency‑domain EEG features with two dimensionality‑reduction stages to improve accuracy and computational efficiency in epilepsy diagnosis. EEG signals are first decomposed using DWT, and informative sub-bands are selected. From these, time-domain features such as sub-pattern energy, spike rhythmicity, and rhythmic spike amplitude, and frequency-domain features such as dominant frequency and spectral entropy from a 32-point Fast Fourier Transform (FFT) are extracted. Additional wavelet-based indices provide time-frequency descriptors, with energy identified as the most discriminative. To reduce feature dimensionality, two PCA-based methods are applied. In the first method, each feature vector is divided into sub-parts, PCA is applied to each part, and the leading components are concatenated to form a compact global feature vector. The second method further compresses this output by projecting the data onto a selected set of principal components. Both approaches reduce computational complexity while maintaining or improving classification accuracy. The extracted features are then classified using models including SVM, KNN, RF, MLP, LightGBM, and CatBoost.
	
	Savadkoohi et al.~\cite{ref82} proposed a streamlined machine learning framework for epileptic seizure detection that emphasizes precise feature engineering and dimensionality reduction. To capture the dynamic characteristics of EEG signals, the authors extracted four statistical metrics—mean, variance, kurtosis, and skewness—across three distinct analytical domains. This was achieved by applying a third-order Butterworth filter in the Time Domain (TD), a FFT in the Frequency Domain (FD), and a DWT in the Time-Frequency Domain (TFD). By evaluating these metrics across the five primary brain frequency bands, an initial feature space of sixty variables was generated. To optimize classification and reduce computational overhead, this space was refined using a two-stage feature selection strategy consisting of a Student's t-test ($p < 0.05$) followed by sequential forward floating selection. The resulting optimal feature set was then fed into SVM and KNNs classifiers to determine the system's overall diagnostic accuracy.
	
	\subsection{Deep Learning Architectures}
	Recent studies have introduced a range of DL architectures tailored for EEG-based seizure detection:
	
	\begin{itemize}
		\item \textbf{CNNs} – Capture spatial relationships between electrodes and local temporal dependencies.
		\item \textbf{RNNs} and \textbf{LSTM} – Model sequential dependencies in EEG signals over time.
		\item \textbf{Hybrid Models (CNN-LSTM, BiLSTM, or Attention-based networks)} – Combine spatial and temporal learning for improved performance.
	\end{itemize}
	
	These models are advantageous because they can automatically learn complex temporal and spatial patterns directly from EEG data, reducing reliance on handcrafted features. CNNs are highly effective in capturing spatial relationships across multiple electrodes, while RNNs especially LSTMs \cite{ref93} are better suited for representing long-term sequential dependencies. By combining these strengths, hybrid architectures such as CNN-LSTM \cite{ref94} or BiLSTM \cite{ref95} provide a unified framework capable of learning spatial–temporal correlations, leading to improved robustness and generalization in seizure detection tasks. In the following we reviewed several papers for each categories that used DL approaches for seizure detection.  
	
	\subsubsection{Spatial Architectures}
	A novel approach in EEG‑based detection is time‑series imaging. In this method, time–frequency maps can be generated for each electrode \cite{ref52}. These two‑dimensional images contain both spectral and temporal information, allowing the use of 2D CNN models for their classification. In addition to 2D CNNs, 1D CNNs also provide high accuracy for EEG signal classification. One advantage of 1D CNNs over 2D CNNs is that they require fewer parameters during training; therefore, they need less data to achieve convergence \cite{ref72}. It should be noted, however, that 1D CNNs achieve high accuracy only when precise and separate annotations are available for each EEG channel, which leads to more intensive preprocessing during labeling. In cases where annotations are weak, 2D CNN models yield better results, because regardless of which channel the seizure occurs in, only the start and end of the seizure episode need to be labeled \cite{refn6}.
	
	In the following, we review several studies on the use of 2D CNN models for classifying images derived from EEG signals, as well as 1D CNN models for categorizing time‑series EEG signals.
	
	Iešmantas and Alzbutas~\cite{ref8} proposed a seizure detection and classification framework that operates on clinical EEG data. In this study, brain activity images were constructed using various features such as Phase Locking Value (PLV), entropy, and energy. These images enable the use of CNN for classification. A key distinction of this work from much of the existing literature is the use of extremely short EEG segments for analysis. Whereas many prior approaches rely on long temporal windows to capture seizure dynamics, the proposed framework processes EEG data at a fine temporal granularity, enabling improved time localization of seizure events and reducing detection latency. While shorter windows inherently limit the amount of available information and increase feature variability, the adoption of a CNN allows the model to learn discriminative spatial and frequency-dependent patterns across EEG channels, thereby mitigating the limitations associated with short-duration signal segments.
	
	Gómez et~al.~\cite{ref67} proposed an automatic seizure detection approach based on imaged EEG representations and fully convolutional networks. In this method, EEG signals are transformed into two-dimensional images, with rows corresponding to channels, columns to time, and pixel intensity representing signal amplitude. This representation captures both spatial patterns across electrodes and temporal dynamics of the EEG. During preprocessing, minimal intervention is applied, limited to saturating the signal at  $\pm 250~\mu$V to reduce noise, without filtering or amplitude transformations. To address the imbalance between seizure and non-seizure data, overlapping windows are generated for seizure segments. The fully convolutional networks architecture is employed for detection, and the model is trained and evaluated using a LOO strategy along with the first seizure model approach.
	
	Tanveer et~al.~\cite{ref72} proposed a DL framework for neonatal seizure detection using multichannel EEG signals, addressing the challenges posed by subtle seizure patterns and the absence of a post-ictal phase. In the preprocessing stage, a notch filter was applied to remove power-line noise, and the EEG signals were reshaped into a 2D representation. The data were segmented into 1-second windows, with seizure segments overlapped by 50\% to increase training samples while preserving temporal resolution. A CNN was employed for binary classification. To account for variability in expert annotations, three independent CNN models were trained using labels from three clinicians, and the final decision was obtained through consensus among the models.
	
	Sharan and Berkovsky~\cite{ref73} proposed a DL-based framework for epileptic seizure detection from EEG signals that reduces reliance on manual feature engineering by learning directly from spectral representations. EEG recordings were segmented into non-overlapping 2-second windows, and class imbalance was addressed by using all seizure segments and randomly selecting an equal number of non-seizure segments for training and validation. A Butterworth filter was applied during preprocessing, followed by frequency-domain transformation using FFT and continuous wavelet power spectra. The resulting spectral representations were classified using a 1D CNN.
	
	Liu et~al.~\cite{ref75} presented an automated seizure detection framework that integrates time–frequency signal processing with DL for iEEG analysis. In the preprocessing stage, raw multichannel signals are segmented into 4-second windows and filtered using DWT to remove noise and artifacts. The Stockwell transform (S-transform), which combines the advantages of STFT and continuous wavelet transform, is applied to capture non-stationary characteristics and generate time–frequency spectrograms. These spectrograms are converted into image-like representations suitable for CNN input. A 15-layer CNN is then employed for automatic feature extraction and classification. Finally, post-processing steps, including smoothing, thresholding, and collar operations, are applied to enhance detection accuracy and reduce false alarms.
	
	Wang et~al.~\cite{ref77} proposed an automatic seizure detection and multi-class classification framework that combines a 1D CNN with moth flame optimization for automated hyper-parameter tuning. The framework aims to improve early seizure detection while minimizing the need for manual feature engineering and network parameter adjustments. EEG signals were preprocessed using a band-pass filter and segmented into 1-second windows. Leveraging the temporal characteristics of EEG signals, the 1D CNN architecture reduces computational complexity and the number of parameters while effectively learning discriminative features for seizure classification.
	
	One of the main limitations of deep neural network–based EEG analysis is the limited interpretability of model decisions, which restricts clinical adoption. To address this issue, Raab et~al.~\cite{ref52} introduced XAI4EEG as an explainable framework for transparent DL-based EEG diagnosis. In XAI4EEG, multichannel EEG signals are preprocessed to retain relevant spectral, temporal, and spatial information for seizure analysis. The framework combines two complementary models: a 1D CNN to capture temporal and frequency-related patterns within individual channels, and a 3D CNN to model short-range spatio-temporal dependencies across multiple electrodes. This dual-model design enables simultaneous analysis of local channel dynamics and inter-channel spatial interactions. To enhance interpretability, SHAP-based local explanation techniques are employed to compute attribution scores, which are mapped onto clinically meaningful dimensions, including EEG frequency bands, electrode locations, and temporal segments. By explaining predictions in terms of where, when, and in which frequency ranges relevant patterns occur, XAI4EEG provides intuitive and clinically aligned explanations.
	
	The dataset used, the preprocessing steps applied, and the architecture employed in all reviewed studies are summarized in Table~\ref{tab:CNNbased}.
	
	\begin{table}[!htbp]
        \centering
        
        \begin{threeparttable}
            \caption{Summary of CNN-Based EEG Seizure Detection Studies.}
            \label{tab:CNNbased}
            \tiny
            \renewcommand{\arraystretch}{1.15}
            \setlength{\tabcolsep}{3pt}
            
            \begin{tabular}{
                    p{1.0cm} 
                    p{1.4cm} 
                    p{3.8cm} 
                    p{5.0cm} 
                    p{3.2cm} 
                }
                \toprule
                \textbf{Study} &
                \textbf{Dataset} &
                \makecell[l]{\textbf{Preprocessing /}\\ \textbf{Feature Extraction}  \textbf{(Selection)}} &
                \textbf{Classification Architecture} &
                \textbf{Key Results (\%)} \\
                \midrule 
                
                \makecell[l]{Iešmantas \& \\Alzbutas\\ 2020\,\cite{ref8}} &
                \makecell[l]{\hspace{3pt}TUH}  &
                Window segmentation; \newline
                    Band-pass filtering &
                2$\times$[Conv(10, 5$\times$5)] $\rightarrow$ MaxPool(2) \newline $\rightarrow$ FC(1000) &
                \makecell[l]{SEN - SPE - AUC \\
                    68 - 67 - 74} \\ \\ 
                
                \makecell[l]{Gómez \\et al.\\ 2020\,\cite{ref67} }&
                CHB-MIT,\newline EPILEPSIAE &
                \makecell[l]{EEG-to-image trans;\\
                    Channel standardization;\\
                    4-s window segmentation;\\ 
                    saturation at $\pm250\,\mu$V;\\
                } &
                3$\times$[Conv $\rightarrow$ BN $\rightarrow$ ReLU $\rightarrow$ Pool] \newline
                    $\rightarrow$ 2$\times$Conv (FC replacement) \newline
                    $\rightarrow$ SoftMax &
                \makecell[l]{ACC - SPE - PRE - FAR \\
                    $99.3$ - $99.6$ - $62.7$ - $<$0.5/h \tnote{$\dagger$} \\
                    $98.0$ - $98.3$ - $22.9$ - $<$1.0/h \tnote{$\ddagger$} } \\ \\
                
                Tanveer et al.\newline 2021\,\cite{ref72} &
                NICU (Helsinki) &
                50 Hz notch filter;\newline reshape to 19$\times$N; \newline 1-s sliding window (50\% overlap) &
                3$\times$[Conv2D(32, 3$\times$4) $\rightarrow$ ReLU] \newline
                    $\rightarrow$ BN \newline
                    $\rightarrow$ AvgPool(1$\times$8) \newline
                    $\rightarrow$ 2$\times$[Conv2D(32, 3$\times$4) $\rightarrow$ \newline ReLU] \newline
                    $\rightarrow$ AvgPool(1$\times$4) \newline
                    $\rightarrow$ Conv2D(1, 3$\times$4) \newline
                    $\rightarrow$ Global AvgPool \newline
                    $\rightarrow$ FC &
                \makecell[l]{ACC - AUC \\
                    $95.6$ - $99.2$ \tnote{$\mathsection$} \\
                    $94.8$ - $98.4$ \tnote{$\mathparagraph$} \\
                    $90.1$ - $96.7$ \tnote{$\parallel$} } \\ \\
                
                Sharan \&\newline Berkovsky\newline 2020\,\cite{ref36} &
                CHB-MIT &
                \makecell[l]{Butterworth filtering ($f_c = 45\,\text{Hz}$);\\ 2-s window segmetation; WT} &
                3$\times$ [Conv (128 filters, $(K=15,10,5)\times1$, \newline stride $1\times1$) $\rightarrow$ ReLU]
                $\rightarrow$ \newline MaxPool ($2\times1$, stride $2\times1$) \newline
                    $\rightarrow$ Fully Connected $\rightarrow$ Softmax &
                \makecell[l]{ACC - SEN - SPE \\
                    $97.25$ - $97.25$ - $97.25$} \\ \\
                
                Liu et al.\newline 2019\,\cite{ref75} &
                iEEG dataset (Freiburg) &
                4-s window segmentation;\newline
                    DWT filtering (Db4);\newline
                    1-s widow segmentation &
                4$\times$ [Conv (163, 32, 64, 64) $\rightarrow$ BN $\rightarrow$ ReLU] \newline 
                    $\rightarrow$ MaxPool after Conv$_{1,2,4}$ $\rightarrow$ \newline Dropout (20\%) after Conv$_{3,4}$ \newline 
                    $\rightarrow$ FC (120) $\rightarrow$ FC (2) $\rightarrow$ Softmax &
                \makecell[l]{ACC - SEN - SPE - AUC \\
                    $98.12$ - $97.01$ - $98.12$ - $97.21$} \\ \\
                
                \makecell[l]{Kuang \\et al.\\ 2024\,\cite{ref76} }&
                UCI &
                \makecell[l]{Band-pass filtering (0.53--40 Hz);\\ 1-s windows segmentation;\\ resampling(256Hz to 178Hz); \\Z-score normalization }&
                3$\times$[Conv (20 filters, 5$\times$1, stride 1) $\rightarrow$ \newline ReLU  $\rightarrow$ Dropout (0.5)] \newline
                    $\rightarrow$ MaxPool (2$\times$1) \newline 
                    $\rightarrow$  Flatten \newline
                    $\rightarrow$ FC (50, ReLU) \newline
                    $\rightarrow$ FC (1, Sigmoid) &
                \makecell[l]{ACC - SEN - SPE \\
                    $98.70$ - $97.53$ - $98.98$} \\ \\
                
                \makecell[l]{Wang \\et al.\\ 2023\,\cite{ref77}} &
                Bonn  &
                Band-pass filtering(0.53--40 Hz);\newline 1-s window segmentation &
                $6\times$[Conv$\rightarrow$BN$\rightarrow$Dropout (0.2)$\rightarrow$MaxPool]\newline $\rightarrow$Tiling $\rightarrow$ FC $\rightarrow$ 5-class output &
                \makecell[l]{ACC - F1 \\
                    $99.96$ - $99.90$}\\ \\
                    
                \makecell[l]{Raab \\et al.\\ 2022\,\cite{ref52}} &
                NICU (Helsinki) &
                \makecell[l]{Band-pass filtering (0.5--12.5 Hz);\\ 1-s window segmentation; \\
                interval dissection; spectral analysis;\\ interval recomposing (1D) \\
                time-frequency analysis; \\3D image construction (3D) } &
                1D CNN: 3$\times$[Conv(64/128/256; k=3/3/2)\newline
                $\rightarrow$ BN $\rightarrow$ Dropout(0.2) $\rightarrow$ Pooling]\newline
                $\rightarrow$ Flatten $\rightarrow$ FC(16) $\rightarrow$ FC(1) &
                \makecell[l]{ACC - SEN - SPE - PRE\\
                    $90.06$ - $82.57$ - $97.55$ - $84.24$ \tnote{$\sharp$} \\
                    $89.07$ - $86.00$ - $92.14$ - $63.04$ \tnote{$\star$} } \\
                
                \bottomrule
            \end{tabular}
            
            \begin{tablenotes}\footnotesize
                \item[$\dagger$] Results correspond to the CHB-MIT dataset.
                \item[$\ddagger$] Results correspond to the EPILEPSIAE dataset.
                \item[$\mathsection$] Results evaluated on Expert 1 annotations.
                \item[$\mathparagraph$] Results evaluated on Expert 2 annotations.
                \item[$\parallel$] Results evaluated on Expert 3 annotations.
                \item[$\sharp$] Results obtained using 1D CNN architecture.
                \item[$\star$] Results obtained using 3D CNN architecture.
            \end{tablenotes}
        \end{threeparttable}
        
    \end{table}
	
	\subsubsection{Temporal Architectures}
	RNN-based methods have been widely adopted in epileptic seizure detection due to their ability to model temporal dependencies and sequential patterns in EEG signals. One of the key advantages of RNN architectures is their ability to automatically learn discriminative features directly from time-series data. Several variants of RNNs have been developed, including LSTM and BiLSTM, which are particularly effective for modeling temporal dependencies. In recent studies, RNN-based frameworks have frequently been combined with advanced preprocessing and feature extraction techniques to enhance performance and achieve more accurate seizure detection.
	
	Hu et~al.~\cite{ref79} proposed a seizure detection framework that combines LMD with a BiLSTM network for accurate analysis of long-term EEG recordings. In the preprocessing stage, LMD is applied to decompose the EEG signals into multiple Product Functions (PFs) and a baseline component. The first three PFs, which contain the most relevant seizure-related information, are selected for further analysis. From each PF, ten statistical features are extracted and combined into a single feature vector. The preprocessing procedure is computationally efficient (approximately 0.015 seconds per segment), enabling potential real-time implementation. The extracted feature vectors are then classified using a BiLSTM network to capture temporal dependencies in both forward and backward directions. In the post-processing stage, a central linear moving average filter is applied to smooth the model outputs and reduce false positives, followed by a thresholding strategy to generate final binary seizure/non-seizure labels.
	
	 Tang et~al.~\cite{ref80} proposed an automatic epilepsy detection framework that integrates path signature feature extraction with a BiLSTM network and an attention mechanism. EEG signals are first denoised using an FIR filter, segmented into fixed-length windows, and converted to a bipolar bipolar temporal central parasagittal  montage for the TUEP dataset. The path signature algorithm captures dynamic inter-channel relationships as features, which are classified by the BiLSTM network, while the attention layer enhances important feature selection.
	
	 Singh and Malhotra~\cite{ref81} presented a DL framework for automatic epileptic seizure prediction using long-term multichannel EEG signals.
	 The approach aims to improve early detection of (pre-seizure) states by extracting frequency-domain features from EEG data and modeling their temporal patterns using RNN. In the preprocessing stage, a Butterworth band-pass filter was applied to remove noise and artifacts. Because EEG signals are inherently non-stationary, the recordings were segmented into windows ranging from 5 to 50 seconds to achieve quasi-stationary behavior, ensuring more consistent statistical properties and reduced computational complexity. The segmented signals were then transformed into the frequency domain using the FFT, and two spectral features (spectral power and mean spectral amplitude) were extracted from five standard frequency bands. These features were used as input to LSTM-based classifiers. Two architectures were evaluated: a single-layer LSTM (1L-LSTM) and a two-layer LSTM (2L-LSTM). Experimental results demonstrated that the 2L-LSTM model outperformed both the 1L-LSTM and traditional ML classifiers.
	All these reviewed papers are inserted briefly in Table~\ref{tab:RNNbased}.
	
	\begin{table}[!htbp]
        \centering
        
        \begin{threeparttable}
            \caption{Summary of RNN–Based EEG Seizure Detection Studies.}
            \label{tab:RNNbased}
            \tiny 
            \renewcommand{\arraystretch}{1.2}
            \setlength{\tabcolsep}{3pt}
            
            \begin{tabular}{
                    p{1.2cm} 
                    p{1.3cm} 
                    p{4.0cm} 
                    p{4.0cm} 
                    p{4.0cm} 
                }
                \toprule
                \textbf{Study} & \textbf{Dataset} & \makecell[l]{\textbf{Preprocessing /}\\ \textbf{Feature Extraction}  \textbf{(Selection)}} & \makecell[l]{\textbf{Classification}  \textbf{Architecture}} & \textbf{Key Results (\%)} \\
                \midrule
                
                Hu et al.\newline 2020\,\cite{ref79} & 
                CHB-MIT & 
                \makecell[l]{4-s sliding windows;\\ LMD decomposition\\ (PF1--PF3); \\ statistical features} & 
                BiLSTM (64 units, dropout=0.35) $\rightarrow$\newline
                    FC + Softmax $\rightarrow$\newline MAF smoothing + thresholding & 
                \makecell[l]{SEN - SPE \\
                    93.61 - 91.85} \\ \\
                
                \makecell[l]{ Tang et al.\\ 2024\,\cite{ref80} }& 
                \makecell[l]{CHB-MIT,\\ TUH, \\ Private EEG} & 
                \makecell[l]{FIR filter(0.5–50 Hz);\\ 1-s window segmentation; \\ Path Signature} & 
                BiLSTM (2 layers, 128 units) $\rightarrow$\newline
                    Attention $\rightarrow$ FC + Softmax & 
                \makecell[l]{ACC - SEN - SPE - PRE - F1\\
                    99.09 - 99.28 - 98.95 - 98.53 - 98.89 \tnote{$\dagger$} \\
                    99.87 - 99.82 - 99.91 - 99.91 - 99.87 \tnote{$\ddagger$} \\
                    99.40 - 99.15 - 99.62 - 99.58 - 99.36 \tnote{$\mathsection$}} \\ \\
                
                Singh et al.\newline 2022\,\cite{ref81} & 
                CHB-MIT & 
                \makecell[l]{Butterworth filter (0.1--127 Hz);\\
                    5--50s segmentation;
                    FFT;\\
                    Frequency band separation;  \\ Mean Spectrum Amplitude; \\Spectral Power} & 
                1L-LSTM (50 units) / \newline 2L-LSTM (128, 64 units)  $\rightarrow$\newline
                    Dropout (0.25) $\rightarrow$ FC(16) + Sigmoid & 
                \makecell[l]{ACC - SEN - SPE - F1\\
                    $97.34$ - $96.44$ - $98.25$ - $97.20$ \tnote{$\mathparagraph$} \\
                    $98.14$ - $98.51$ - $97.78$ - $98.23$ \tnote{$\parallel$}} \\
                
                \bottomrule
            \end{tabular}
            
            \begin{tablenotes}\footnotesize
                \item[$\dagger$] Results correspond to the CHB-MIT dataset.
                \item[$\ddagger$] Results correspond to the TUH (256Hz) dataset.
                \item[$\mathsection$] Results correspond to the Private EEG dataset.
                \item[$\mathparagraph$] Results obtained using 1L-LSTM architecture with 50-s segmentation.
                \item[$\parallel$] Results obtained using 2L-LSTM architecture with 30-s segmentation.
            \end{tablenotes}
        \end{threeparttable}
        
    \end{table}
	
	\subsubsection{Hybrid and Spatio-Temporal Architectures}
	Automatic seizure detection using multi-channel EEG faces the critical challenge of simultaneously modeling spatial correlations among electrodes and the temporal evolution of brain activity. Relying on a single network architecture often proves insufficient for this dual requirement. To address this limitation, recent studies have introduced hybrid DL frameworks that combine complementary model capabilities. In particular, convolutional networks are employed to capture spatial structures, while recurrent architectures such as LSTMs are utilized to characterize temporal dependencies. By integrating these strengths, hybrid models provide a more comprehensive representation of the spatio-temporal characteristics of epileptic activity. The following sections highlight advanced approaches that adopt this integrated strategy to achieve improved performance in seizure detection.
	
	Liu et~al.~\cite{ref31} proposed a pseudo-3D convolutional neural network for EEG-based seizure detection to better capture spatio-temporal dependencies in multi-channel recordings.
	Instead of using full 3D convolution, the model decomposes the operation into two stages: a 2D convolution to extract spatial relationships among EEG channels at each time frame, followed by a 1D convolution to model temporal dynamics by tracking the evolution of these spatial features over time. This design preserves the ability to learn complex spatio-temporal patterns while significantly reducing computational cost and parameter size compared to conventional 3D CNNs. Evaluated on the CHB-MIT dataset, the framework outperformed standard CNN and LSTM models, largely due to its efficient pseudo-3D structure combined with attention-guided recurrent modules.
	
	Omar et~al.~\cite{ref5} proposed a hybrid DL framework (LConvNet) that integrates CNNs with LSTM units to capture both spatial and temporal characteristics of EEG signals. Designed for efficient operation in clinical settings with limited computational resources, EEG recordings were segmented into short overlapping windows, normalized, and reduced in dimensionality. The convolutional block used progressively larger filters (16, 32, 64) and kernels ($3 \times 3$, $5 \times 5$, $7 \times 7$) to extract hierarchical spatial features. A time-distributed layer preserved temporal dynamics before the LSTM module learned long-term dependencies across windows. Global average pooling and dense layers then consolidated representations for binary classification. The model achieved a balance of simplicity and representational power, delivering 97\% accuracy on the TUH EEG Epilepsy Corpus, outperforming EEGNet (86\%) and ShallowConvNet (78\%). When combined with scaled custom attention, accuracy increased to 98.17\%, demonstrating the benefit of attention mechanisms for temporal feature learning.
	
	Liu et~al.~\cite{ref30} proposed a hybrid CNN-BiLSTM model to extract temporal and spatial EEG features, similar to Omar et~al.~\cite{ref5}. Unlike Omar et~al.~\cite{ref5}, which focused on classification accuracy, this study aims to reduce patient-specific bias and improve generalization across individuals. To achieve this, a Channel-Perturbation (CP) training strategy injects controlled noise into input channels, forcing the network to learn features common across patients rather than subject-specific patterns.
	
	The framework combines convolutional layers for spatial feature extraction with bidirectional LSTM layers for temporal modeling, capturing dependencies from both past and future brain activity. EEG signals are preprocessed using the DWT before input to the network. Experiments show that the CP-enhanced model reliably detects seizures across multiple patients, making it well suited for clinical applications where individualized EEG data may be limited.
	
	El-Dajani et~al.~\cite{ref32} developed a patient-independent DL framework for epileptic seizure detection, addressing two major limitations observed in previous research, including the reliance on a large number of EEG electrodes and the need for patient-specific training. The proposed method is designed to function effectively with fewer EEG channels, which makes it suitable for portable and real-time monitoring systems. To capture the temporal dependencies in EEG signals, three RNN architectures are investigated. The first model employs a single-channel LSTM network that builds upon earlier work by Hussein et~al.~\cite{ref37}. This architecture incorporates an LSTM layer for temporal feature extraction, a dropout layer to reduce overfitting, a fully connected layer, a global average pooling layer, and a sigmoid activation function for binary classification.
	The multi-channel LSTM network is built upon the general architecture of the single-channel model and is adapted to process multiple EEG channels concurrently. In this design, each EEG channel is treated as an independent feature source, which may reduce efficiency when the number of channels increases. In addition to these two architectures, a hybrid CNN–LSTM framework is also investigated, combining the spatial feature extraction capability of CNN layers with the temporal modeling strength of LSTM. The CNN component consists of three consecutive 2D CNN layers with 16, 32, and 64 kernels, respectively, each followed by a max pooling layer to progressively reduce dimensionality and computational complexity. The extracted spatial feature maps are then passed to a bidirectional LSTM module, as described by Golmohammadi et~al.~\cite{ref38}, allowing the model to capture temporal dependencies in both forward and backward directions. Experimental results demonstrate that the hybrid CNN–LSTM architecture achieves higher detection performance compared to the standalone LSTM models.
	
	Cao et~al.~\cite{ref48} improved seizure detection accuracy using a deep learning framework combined with feature fusion. EEG signals were first denoised using a Butterworth bandpass filter and then decomposed into five levels via DWT to extract multi-resolution frequency information. To effectively capture the nonlinear and nonstationary characteristics of EEG signals, an entropy fusion strategy incorporating fuzzy membership functions was employed, enhancing robustness to noise and parameter variations. Subsequently, an SVM-based recursive feature elimination method reduced the feature set from 24 to 12 salient features, which were finally classified using a hybrid CNN–BiLSTM model to exploit both discriminative patterns and temporal dependencies.
	
	Guhdar et~al.~\cite{ref51} showed that a hybrid framework integrating one-dimensional CNNs with wavelet-based signal decomposition and a multi-head attention mechanism can significantly enhance recognition accuracy.
		
	Awais et~al.~\cite{ref69} introduced a graph-based framework for EEG seizure detection. EEG signals are segmented into 30-s windows with 30\% overlap, and class imbalance is addressed using SMOTE and KNNOR oversampling to generate synthetic seizure samples. Each window is modeled as a graph node, with edges defined by Euclidean similarity between feature vectors, enabling the capture of inter-window and inter-feature dependencies. Two hybrid models are explored; GCN-LSTM, which combines graph convolution for relational feature learning with LSTM-based temporal modeling, and GCN-BRF, where GCN-extracted features are classified using a balanced random forest to improve robustness against class imbalance.
	
	Nogay and Adeli~\cite{ref70} proposed a DL–based framework for automatic epileptic seizure detection and classification using EEG signals and pre-trained AlexNet models. In this approach, one-dimensional EEG recordings are transformed into two-dimensional spectrogram images to capture both temporal and frequency-domain characteristics of the signals. To enhance dataset size and improve model generalization, several data augmentation techniques such as horizontal shifting, rotation, brightness adjustment, zooming, and cropping are applied. The study adopts a transfer learning strategy by fine-tuning AlexNet, in which the last three layers are replaced to accommodate both binary and ternary classification tasks. In addition, entanglement learning is employed to further improve seizure classification performance.
	
	Najafi et~al.~\cite{ref71} proposed an EEG-based framework for seizure and epilepsy diagnosis. First, EEG signals are classified as normal or epileptic, followed by the identification of focal or generalized epilepsy. During preprocessing, DWT and Longitudinal Bipolar montage (LB) are applied, and signals are segmented into 1-second windows with 50\% overlap. From each segment, statistical, root-based, and spectral power features across standard EEG bands are extracted. Feature selection uses pearson correlation and hypothesis testing, with low-correlation features grouped and highly correlated features treated separately to ensure informative, non-redundant inputs. 
	
	Abdelhameed and Bayoumi~\cite{ref74} proposed a DL framework for automatic epileptic seizure detection that learns discriminative patterns directly from EEG signals using autoencoder-based architectures. A two-dimension deep convolutional autoencoder (2D-DCAE) is employed for automatic feature extraction, where the encoder compresses multi-channel EEG data into a low-dimensional latent representation and the decoder reconstructs the original signal. Based on the learned features, two classification frameworks are introduced: 2D-DCAE + MLP and 2D-DCAE + BiLSTM, enabling both spatial feature learning and temporal pattern modeling for seizure detection.
	
	Zhao et~al.~\cite{ref78} proposed a hybrid deep learning framework, ResBiLSTM, that integrates one-dimensional residual networks (ResNet) for spatial feature extraction with BiLSTM networks for temporal sequence modeling in EEG-based seizure detection. Data preprocessing included segmentation into shorter windows and noise-based data augmentation to increase training samples and improve robustness. ResBiLSTM was compared with recent RNN-, CNN-, and hybrid-based methods (e.g., CE-stSENet, VWCNNs, and 3D-CBAMNet). The results showed superior performance across binary, three-class, and five-class tasks, strong generalization on both single- and multi-channel datasets, and robust handling of imbalanced data, reflected by high weighted F1 scores on TUSZ. 
	All these reviewed papers are inserted briefly in Table~\ref{tab:Hybridbased_Seizure}.
	
	\begin{table}[!htbp]
        \centering
        
        \begin{threeparttable}
            \caption{Summary of Hybrid Architecture–Based EEG Seizure Detection Studies.}
            \label{tab:Hybridbased_Seizure} 
            \tiny 
            \renewcommand{\arraystretch}{1.2}
            \setlength{\tabcolsep}{3pt}
            
            \begin{tabular}{
                    p{1.3cm} 
                    p{1.5cm} 
                    p{4.3cm} 
                    p{4.2cm} 
                    p{3.6cm} 
                }
                \toprule
                \textbf{Study} & \textbf{Dataset} & \makecell[l]{\textbf{Preprocessing /}\\ \textbf{Feature Extraction (Selection)}} & \textbf{Classification Architecture} & \textbf{Key Results (\%)} \\
                \midrule
                
                Liu et al.\newline 2024~\cite{ref31} & 
                \makecell[l]{CHB-MIT} & 
                \makecell[l]{Band-pass filtering (0.5–75\,Hz); \\ ICA; Segmentation; \\ HFD, ApEn, SampEn, FuzzyEn;\\mRMR }& 
                Pseudo-3D CNN $\rightarrow$\newline BiConvLSTM3D $\rightarrow$\newline 3D Attention & 
                \makecell[l]{ACC - SEN - SPE - PRE \\
                    98.13 - 98.03 - 98.23 - 98.30} \\ \\
                    
                Omar et al.\newline 2023~\cite{ref5} & 
                TUH & 
                resampled to 128\,Hz;\newline
                Band-pass filtering; ICA\newline Segmentation\,s;\newline
                2-s sliding window (50\% overlap);\newline
                PCA (25 components, $>$90\% CEVR) & 
                CNN $\rightarrow$\newline TimeDistributed $\rightarrow$ LSTM & 
                \makecell[l]{ACC - PRE - F1 \\
                    97 - 97 - 97} \\ \\   
                
                Liu et al.\newline 2021~\cite{ref30} & 
                \makecell[l]{CHB-MIT} & 
                4-s window segmentation;\newline resampled to 256\,Hz \newline(FIR antialiasing); \newline
                    DWT (Db4) & 
                1D CNN $\rightarrow$ BiLSTM & 
                \makecell[l]{ACC - SEN - SPE \\
                    97.51 - 83.11 - 97.58} \\ \\
                
                \makecell[l]{El-Dajani \\et al.\\ 2025~\cite{ref32}} & 
                Bonn & 
                \makecell[l]{Segmentation; under-sampling;\\ over-sampling; Sliding window \\(90\% overlap); Channel reduction} & 
                \makecell[l]{Single-channel LSTM;\\ Multi-channel LSTM;\\ 3$\times$2D CNN $\rightarrow$ BiLSTM} & 
                \makecell[l]{ACC - SEN - AUC \\
                    100 - 74 - 92 \tnote{$\dagger$} \\
                    100 - 55 - 90 \tnote{$\ddagger$} \\
                    100 - 68 - 92 \tnote{$\mathsection$}} \\ \\
                
                Cao et al.\newline 2023~\cite{ref48} & 
                CHB-MIT, Bonn & 
                \makecell[l]{Butterworth filtering(0.53–40\,Hz);\\ DWT (5 levels decomposition);\\ ApEn, FuEn, RMS, Hurst;\\SVM-RFE }& 
                CNN $\rightarrow$ BiLSTM\newline with feature fusion & 
                \makecell[l]{ACC - SEN - SPE - PRE \\
                    98.43 - 97.84 - 99.21 - 99.14 \tnote{$\mathparagraph$} \\
                    100 - 100 - 100 - 100 \tnote{$\parallel$}} \\ \\
                
                \makecell[l]{Guhdar \\et al.\\ 2025~\cite{ref51}} & 
                UCI & 
                Window segmentation;\newline WT; Label adjustment & 
                1D CNN $\rightarrow$\newline Multi-Head Attention $\rightarrow$\newline Fully Connected & 
                \makecell[l]{ACC - F1 \\
                    99.83 - 99.90} \\ \\
                
                Awais et al.\newline 2024~\cite{ref69} & 
                CHB-MIT\newline TUH & 
                \makecell[l]{30-s sliding windows (30\% overlap);\\ oversampling (SMOTE \& KNNOR); \\
                Statistical; frequency-domain;\\ DWT (db8) features } & 
                GCN $\rightarrow$ LSTM;\newline GCN $\rightarrow$ Balanced RF & 
                \makecell[l]{ACC - SEN - SPE \\
                    99.61 - 98.86 - 98.00 \tnote{$\sharp$} \\
                    99.73 - 98.65 - 98.00 \tnote{$\star$} \\
                    99.01 - 98.69 - 98.14 \tnote{$\diamond$} \\
                    99.40 - 98.10 - 97.25 \tnote{$\triangle$}} \\ \\
                
                \makecell[l]{Nogay \& \\Adeli\\ 2021~\cite{ref70}} & 
                Bonn & 
                EEG to spectrogram images;\newline data augmentation(horizontal shift,\newline rotation, brightness, zoom, cropping, \newline image resizing) & 
                Pretrained AlexNet (2D CNN):\newline 5 Conv + 3 MaxPool + ReLU,\newline 3 FC + Softmax \newline with transfer learning & 
                \makecell[l]{ACC - SEN - SPE \\
                    100 - 100 - 100} \\ \\
                
                Najafi et al.\newline 2022~\cite{ref71} & 
                HCTM & 
                \makecell[l]{DWT (Coif3); \\longitudinal bipolar montage; \\1-s sliding window (50\% overlap);\\
                Statistical features (mean, STD, P2P, \\min, max, skewness, kurtosis, P2RMS, \\RSS) and frequency-band powers} & 
                \makecell[l]{RNN-LSTM: Sequence input\\ $\rightarrow$ BiLSTM (200) $\rightarrow$ FC $\rightarrow$ SoftMax }& 
                \makecell[l]{ACC - SEN - SPE \\
                    96.1 - 96.8 - 97.4} \\ \\
                
                \makecell[l]{Abdelhameed\\ \& Bayoumi\\ 2021~\cite{ref74}} & 
                \makecell[l]{CHB-MIT} & 
                \makecell[l]{Z-score normalization; \\
                             Min–Max normalization;\\
                             Input reshaping;\\
                    (1-s, 2-s, 4-s) window segmentation} & 
                2D-DCAE $\rightarrow$ MLP;\newline 2D-DCAE $\rightarrow$ BiLSTM & 
                \makecell[l]{ACC - SEN \\
                    $98.79$ - $98.72$ \tnote{$\nabla$} \\
                    $98.42$ - $98.19$ \tnote{$\infty$}} \\ \\
                
                Wang et al.\newline 2024~\cite{ref78} & 
                \makecell[l]{Bonn}& 
                \makecell[l]{Window segmentation;\\
                     noise-based data augmentation}& 
                ResNet (3$\times$ 1D residual blocks:\newline
                    2$\times$(Conv1D+BN+ReLU) + skip\newline
                    connection; kernel=5$\times$1;\newline
                    dropout=0.2) $\rightarrow$\newline
                    BiLSTM (64 units) $\rightarrow$\newline
                    Dropout(0.2) $\rightarrow$ FC(128)+ReLU $\rightarrow$\newline
                    Dropout(0.5) $\rightarrow$ FC + Softmax & 
                \makecell[l]{ACC - PRE - F1 \\
                    98.88 - 100 - 99.87} \\
                
                \bottomrule
            \end{tabular}
        
            \begin{tablenotes}\footnotesize
                \item[$\dagger$] Results obtained using Single-channel LSTM architecture.
                \item[$\ddagger$] Results obtained using Multi-channel LSTM architecture.
                \item[$\mathsection$] Results obtained using 3$\times$2D CNN $\rightarrow$ BiLSTM architecture.
                \item[$\mathparagraph$] Results correspond to the CHB-MIT dataset.
                \item[$\parallel$] Results correspond to the Bonn dataset.
                \item[$\sharp$] Results correspond to CHB-MIT evaluated with GCN $\rightarrow$ Balanced RF.
                \item[$\star$] Results correspond to CHB-MIT evaluated with GCN $\rightarrow$ LSTM.
                \item[$\diamond$] Results correspond to TUH evaluated with GCN $\rightarrow$ Balanced RF.
                \item[$\triangle$] Results correspond to TUH evaluated with GCN $\rightarrow$ LSTM.
                \item[$\nabla$] Results obtained using 2D-DCAE $\rightarrow$ BiLSTM architecture.
                \item[$\infty$] Results obtained using 2D-DCAE $\rightarrow$ MLP architecture.
            \end{tablenotes}
        \end{threeparttable}
        
    \end{table}

	\section{Prevailing Challenges and Bottlenecks in DL-Based Seizure Detection}
	\label{sec:challenges}
	
	Despite impressive advances in deep learning for EEG-based seizure detection, several persistent challenges restrict its reliable deployment in clinical environments. Based on recent literature, these bottlenecks can be categorized into three dimensions, as illustrated in Table~\ref{tab:challenge_summary}.
	
	\subsection{The Black-Box Dilemma: Interpretability and Clinical Trust}
	
	The most critical barrier to clinical adoption is the inherent opaque nature of conventional DL architectures \cite{ref23, ref61}. While standard CNNs, RNNs, and Transformers achieve high accuracy, they learn highly entangled latent representations that lack transparent functional forms. Consequently, their internal decision boundaries cannot be easily mapped to neurophysiologically meaningful quantities.  While some studies apply post hoc explainable AI (xAI) techniques such as Shapley values or attention heatmaps to identify important channels or time intervals \cite{ref40, ref52, ref69}. These methods typically function as external interpretability layers rather than addressing the inherent opacity of the models. True clinical interpretability requires models to align directly with recognized neurophysiological patterns (e.g., specific frequency-band power or focal ictal dynamics) in a clinician-understandable format \cite{ref23, ref52}. Furthermore, standard architectures offer no straightforward mechanism to mathematically inject domain knowledge such as known patient‑specific rhythms or seizure onset zones into the model’s core structure.
	
	These interpretability shortcomings severely impact clinical usability. The primary goal of automated seizure detection is to serve as a reliable tool that supports clinical decision‑making\cite{ref8, ref25}. Physicians cannot safely rely on algorithms if they cannot assess how they reach their decisions. When false positives occur or subtle seizures are missed, clinicians need to seamlessly debug the error, understand the failure mode, and adjust thresholds \cite{ref6, ref61}. Black-box architectures that provide only binary outputs or abstract heatmaps limit this process, leaving a wide gap between high performance on research benchmarks and the ability to integrate these models into real hospital workflows \cite{ref39, ref69}.
	
	\subsection{Vulnerability in Real-World EEG: Generalization, Robustness, and Data Hunger}
	
	Real-world clinical EEG is widely known to be heterogeneous, noisy, and nonstationary, which reveals the limitations and vulnerability of conventional deep learning models.  Models trained on carefully curated cohorts often experience significant decreases in accuracy when evaluated on new patients or data from external hospital centers \cite{ref23, ref69}. Standard networks often overfit to the particular spatial patterns or baseline rhythms present in their training data distributions \cite{ref40, ref61}. While broad generalization is important, clinical practice also requires personalized approaches. Since seizure patterns vary greatly between individuals, an Productive system must be able to efficiently fine-tune to the specific characteristics of each patient. However, standard deep learning models typically require time-consuming full-network retraining to adapt to new data, lacking a principled and efficient way to fine-tune personalized decision boundaries \cite{ref23, ref25}.
	
	This vulnerability is further exacerbated by the constant presence of artifacts and distribution shifts in clinical settings. Benchmark datasets are often heavily pruned of noise, such as muscle activity, electrode pops, or patient movement \cite{ref39}. In contrast, continuous clinical EEG is riddled with distortions caused by varying medication states, sleep stages, or hardware changes \cite{ref6, ref8, ref30}. Deep learning models that are optimized using clean and well-controlled datasets often demonstrate limited robustness when subjected to the variability, noise, and artifacts characteristic of real-world conditions \cite{ref25, ref32}. Consequently, robustness must be an intrinsic architectural property, rather than depending on aggressive preprocessing steps that may inadvertently remove subtle ictal signals.
	
	Compounding these issues of generalization and robustness is the fundamental challenge of data scarcity and extreme class imbalance. Continuous EEG streams contain only a minute fraction of ictal events compared to normal background activity \cite{ref39, ref61}. High-capacity DL models are exceptionally data-hungry and prone to memorizing background noise when starved of diverse seizure examples \cite{ref23}. Because large-scale, expert‑annotated EEG datasets require enormous effort and time to produce, modern architectures must be designed to learn meaningful representations from limited, imbalanced data without relying heavily on artificial resampling tricks that distort clinical realities \cite{ref25, ref69}.
	
	\subsection{Resource Intensity: Parameter Bloat and Computational Overhead}
	
	Beyond statistical performance, practical seizure detection systems must adhere to strict hardware constraints. Many state-of-the-art DL models, particularly deep Transformers and massive CNN ensembles, are parameter-heavy and computationally expensive \cite{ref40, ref69}. This computational overhead prevents  deployment in resource-constrained environments, such as ultra-low-power implantable devices, ambulatory wearable, or standard bedside monitors, which require real-time, low-latency inference \cite{ref23, ref61}. 
	
	Moreover, the heavy computational overhead of training and hyperparameter tuning makes continuously updated clinical models impractically slow and expensive \cite{ref25}. To bridge the gap between theoretical research and real‑world clinical deployment, the field urgently needs model architectures that preserve the nonlinear expressive power of deep learning while using fewer parameters and significantly less computational effort.
	
	\begin{table}[!htbp]
	    \caption{Summary of DL-Based EEG Seizure Detection Studies and Addressed Challenges Categorized into Three Primary Dimensions.}
	    \label{tab:challenge_summary}
	    \tiny
	    \renewcommand{\arraystretch}{1.3}
	    \setlength{\tabcolsep}{4pt}
	    \centering
	    
	    \begin{tabular}{
	            p{1.7cm} 
	            p{1.5cm} 
	            c        
	            c        
	            c        
	            p{4.5cm} 
	        }
	        \toprule
	        \textbf{Study} &
	        \textbf{Focus} &
	        \parbox{2cm}{\centering\textbf{Interpretability \& Clinical Trust}} &
	        \parbox{2.2cm}{\centering\textbf{Generalization \& Robustness \& Data Hunger}} &
	        \parbox{2.2cm}{\centering\textbf{Parameter Bloat \& Computational Overhead}} &
	        \textbf{Main Contributions Relevant to Challenges} \\
	        \midrule
	        
	        Raab et al. \newline 2023 \cite{ref52} &
	        EEG seizure det.; xAI for clinicians &
	        \checkmark &
	        &
	        &
	        Introduces xAI framework providing spectral, spatial, and temporal visualizations; evaluates clinician trust and understanding; targets clinician‑understandable interpretability. \\
	        \addlinespace
	         
	        Liu et al. \newline 2022 \cite{ref30}&
	        EEG seizure det.; DWT + DL &
	        &
	        \checkmark &
	        &
	        Emphasizes robustness via artifact removal (DWT) and channel-perturbation experiments; discusses intra-patient variability and generalization. \\
	        \addlinespace
	         
	        El‑Dajani et al. \newline 2025 \cite{ref32}&
	        EEG seizure det.; noise robustness &
	        &
	        \checkmark &
	        &
	        Quantifies robustness by injecting noise; compares artifact removal methods (CCA); focuses on patient-independent robustness. \\
	        \addlinespace
	        
	        Y. Sun et al. \newline 2022 \cite{ref40}&
	        Transformer-based iEEG seizure det. &
	        \checkmark &
	        \checkmark &
	        \checkmark &
	        Uses self-attention for explainability; supports patient-independent training; employs knowledge distillation; evaluates on long-term clinical iEEG. \\
	        \addlinespace
	        
	        M. Awais et al. \newline 2024 \cite{ref69}&
	        Graph/Transformer; large clinical data &
	        \checkmark &
	        \checkmark &
	        \checkmark &
	        Uses Shapley values for attribution; improves generalization through Transformers and augmentation; models noise via GCN-LSTM; reduces complexity with channel selection. \\
	        \addlinespace
	 
	        M. K. Siddiqui et al. \newline 2020 \cite{ref6}&
	        Robust EEG artifact handling &
	        \checkmark &
	        \checkmark &
	        &
	        Enumerates clinical artifacts; proposes robust L1-regression and line length; focuses on robustness for noisy clinical EEG. \\
	        \addlinespace
	 
	        Iešmantas \& \newline R. Alzbutas \newline 2020 \cite{ref8}&
	        TUH and clinical corpora &
	        \checkmark &
	        \checkmark &
	        &
	        Compares clean-lab vs noisy-clinical EEG; highlights robustness gap and modest performance on TUH. \\
	        \addlinespace
	
	        E. Ali et al. \newline 2024 \cite{ref25}&
	        Real‑world pediatric EEG &
	        \checkmark &
	        \checkmark &
	        \checkmark &
	        Addresses class imbalance; avoids heavy artifact removal; targets clinically deployable models; optimizes parameters efficiently. \\
	        \addlinespace
	        
	        \bottomrule
	    \end{tabular}
	\end{table}

	\section{An Emerging Paradigm for DL-Based Seizure Detection}
	\label{sec:kans}
	
	The persistent challenges outlined in Section~\ref{sec:challenges}, ranging from black-box opacity and computational overhead to limited generalization, highlight the fundamental limitations of relying solely on conventional deep learning architectures. Recently, Kolmogorov-Arnold Networks have emerged as a powerful new paradigm that directly addresses many of these bottlenecks through a profound structural shift. 
	
	Unlike traditional deep learning models that often require post-hoc patches or complex architectural additions to overcome these limitations, KANs offer a unified mathematical foundation that intrinsically targets several barriers simultaneously. This section shows that, KANs introduce a compact architecture where functions are highly transparent, thereby naturally improving interpretability, mitigating data scarcity, and offering the localized flexibility needed to handle nonstationary EEG artifacts \cite{refk-2,refk-13}.
	
	To understand KANs, it is helpful to compare them with standard MLPs. MLPs are traditionally built upon the Universal Approximation Theorem (UAT) and operate by applying fixed, non-linear activation functions (like ReLU or sigmoid) at every node (or neuron), while the connections (edges) between nodes simply carry static numerical weights. KANs, however, completely invert this design by removing the fixed activation functions from the nodes and instead placing learnable functions on the edges. The nodes then simply sum the incoming signals. Transitioning from static weights to dynamic, trainable edge functions provides a promising mix of high approximation accuracy, improved interpretability, and parameter efficiency for clinical EEG analysis \cite{refk-1,refk-4}.
	
	\subsection{Mathematical Foundations: Representation and Functional Edges}
	
	The mathematical formulation of KANs directly addresses the absence of transparent functional forms and the need for models that gracefully handle non-linear dynamics. 
	
	Instead of the UAT, KANs are explicitly derived from the Kolmogorov-Arnold representation theorem. This theorem is a fundamental result in functional analysis which proves that any complex, multi-variable continuous function $f(x_1,\dots,x_n)$ can be perfectly broken down (decomposed) into a combination of simpler, single-variable (univariate) functions and basic addition \cite{refk-1,refk-4}:
	$$
	f(x_1,\dots,x_n) = \sum_{q=1}^{2n+1} \Phi_q\Bigg( \sum_{p=1}^{n} \phi_{q,p}(x_p) \Bigg)
	$$
	In this equation, each inner function $\phi_{q,p}$ transforms a single input variable, and each outer function $\Phi_q$ aggregates those transformations. For decades, this theorem was considered a theoretical curiosity because the resulting one-dimensional functions were mathematically wild or irregular, making them impossible to train using standard gradient-based machine learning methods \cite{refk-6,refk-5}.
	
	Modern KANs overcome this historical limitation by parameterizing these univariate functions using smooth, well-behaved mathematical tools, most notably B-splines \cite{refk-1,refk-9}. A B-spline can be thought of as a flexible curve made of simple polynomial segments pieced smoothly together. In practice, a KAN layer operates as a matrix of these trainable univariate functions. For an input vector $\mathbf{x} = [x_1,\dots,x_{n_{\text{in}}}]^\top$, the output $\mathbf{y} \in \mathbb{R}^{m}$ of a KAN layer is computed as:
	$$
	y_i = \sum_{j=1}^{n_{\text{in}}} \Phi_{i,j}(x_j), \qquad i=1,\dots,m,
	$$
	where each $\Phi_{i,j}$ is a trainable function rather than a simple numerical weight \cite{refk-1}. To ensure stability during training, these learnable functions are typically split into a basic structural component and a highly flexible spline component:
	$$
	\phi(x) = w_b\, b(x) + w_s \, \text{spline}(x),
	$$
	with
	$$
	\text{spline}(x) = \sum_{k} c_k B_k(x),
	$$
	where $b(x)$ is a fixed basis (e.g., a standard activation like SiLU), $B_k(x)$ represents the B-spline curves, and the coefficients $\{c_k\}$ are what the network actually learns during training \cite{refk-1}. This setup makes the powerful Kolmogorov-Arnold theorem fully compatible with modern backpropagation algorithms.
	
	Because each connection in a KAN is intrinsically expressive, the network achieves global approximation through an efficient ``local-to-global'' mechanism, without requiring very deep architectures \cite{refk-8,refk-9}. Empirically, shallow and relatively compact KANs can match or exceed the predictive performance of significantly larger MLPs, demonstrating their remarkable data efficiency \cite{refk-1,refk-12}. 
	
	Finally, the use of B‑splines equips CNNs with properties that are particularly valuable for biomedical signals such as EEGs. Splines are locally controllable, meaning that adjusting the curve in one specific input range does not influence the function globally \cite{refk-1}. This localized sensitivity is ideal for modeling EEG seizures, where the signal can transition abruptly from stable background noise to  intense, rhythmic, high‑amplitude pathological patterns \cite{refk-2,refk-11}. Furthermore, KANs support grid-extension, a strategy that allows researchers to dynamically increase the resolution of the spline curve to capture finer signal details without needing to erase the model's memory and retrain from scratch \cite{refk-1,refk-9}.

	\subsection{KANs as a Comprehensive Response to the Prevailing Challenges}
	\label{subsec:kan_solution}
	
\begin{figure}[t]
		\centering
		\includegraphics[width=\textwidth]{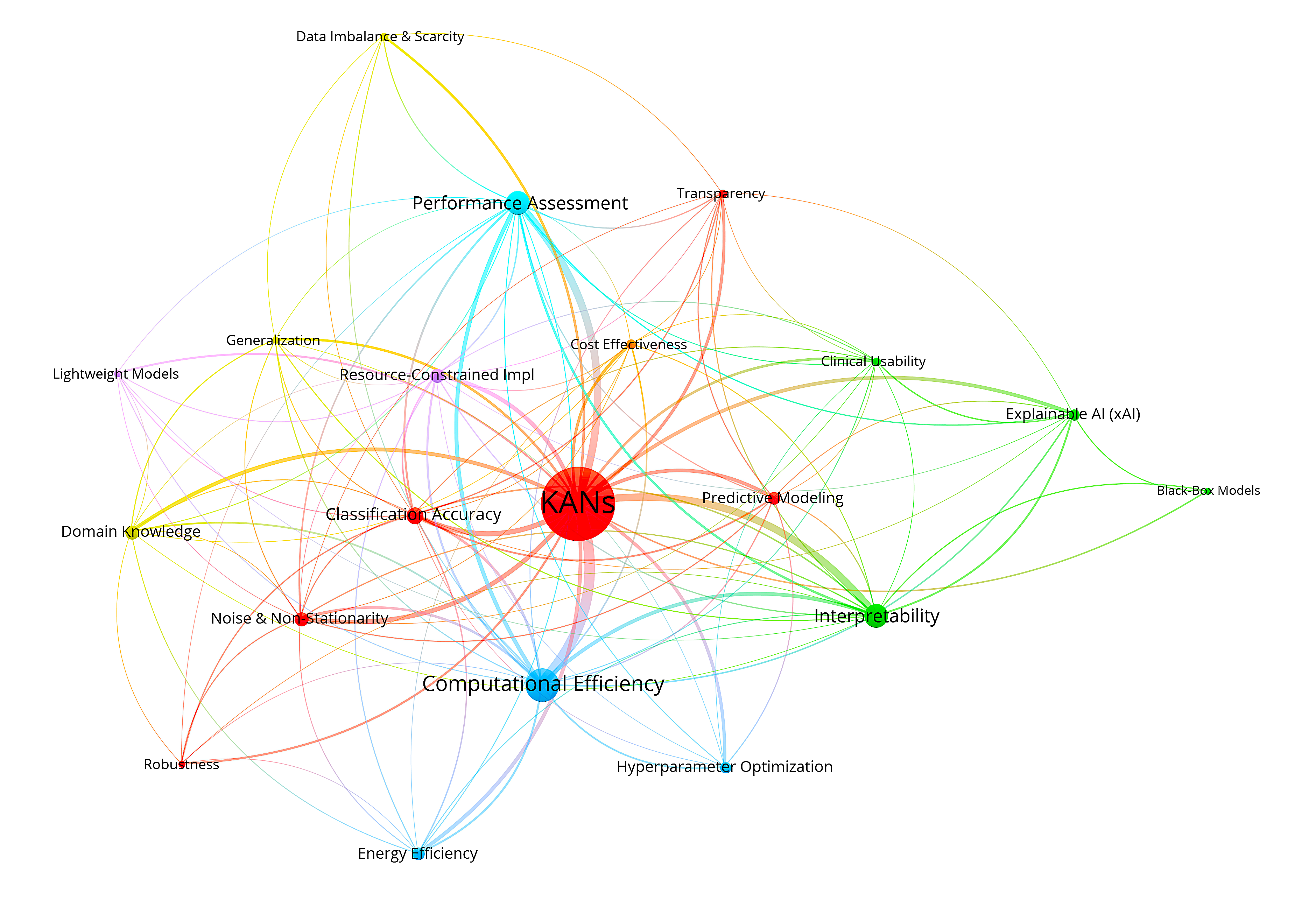} 
		\caption{A conceptual network map illustrating the interconnected relationships between KANs and the prevailing challenges in EEG-based seizure detection, such as interpretability, computational efficiency, and robustness. Data extracted from the Scopus database (May 2026).}
		\label{fig:kan_challenges_network}
	\end{figure}

	The architectural principles of Kolmogorov-Arnold Networks provide a persuasive  and integrated response to the prevailing challenges outlined in Section~\ref{sec:challenges}. Rather than relying on post‑hoc modifications or separate algorithmic solutions for each bottleneck, KANs address these challenges directly through their intrinsic properties. In particular, their learnable edge functions, spline‑based parameterization, and parameter efficiency enable these issues to be handled in a unified manner from the ground up. To illustrate this holistic impact, Figure \ref{fig:kan_challenges_network} presents a conceptual mapping of the relationships between KANs and these prevailing bottlenecks, based on keyword co-occurrence data retrieved from the Scopus database.
	As depicted by the network clusters, the core architecture of KANs inherently branches out to resolve distinct yet interconnected issues, ranging from computational efficiency and data scarcity to robustness and explainable AI.
	
	At the core of this integrated approach is the capacity of KANs to directly resolve the critical challenge of interpretability. Unlike the black-box nature of conventional deep learning models, the functional transparency of KANs allows for the extraction of explicit symbolic equations that describe the learned input-output relationships \cite{refk-16, refk-13, refk-10}. This symbolic regression process converts a trained KAN into a compact, human‑readable mathematical expression. As a result, the model’s decision logic becomes transparent and can be directly related to established neurophysiological knowledge \cite{refk-16}. Consequently, this intrinsic transparency provides clinicians with trustworthy models capable of revealing underlying physical relationships within EEG data, thereby substantially enhancing clinical usability \cite{refk-13}.
	
	Building upon this foundation of transparent clinical utility, the unique topology of KANs establishes a principled framework for personalization. A generalized, cross-patient KAN model can be seamlessly adapted to a new individual by fine-tuning the spline coefficients or by employing grid extension to introduce higher resolution (knots) exclusively where needed for that specific patient's EEG patterns \cite{refk-14}. This adaptability eliminates the need for complete and costly retraining from scratch and aligns with clinical workflows that depend on patient‑specific adjustments. Ultimately, the combination of interpretable insights and localized adaptability positions CNNs as a powerful foundation for developing personalized medicine strategies, effectively bridging the gap between advanced machine learning and practical clinical usage \cite{refk-13}.
	
	The use of B‑splines as the foundation for learnable activation functions plays a key role in making the model more adaptable to individual patients while also improving its overall reliability. This approach naturally enhances the model’s robustness and supports stronger cross‑patient generalization. Because splines are locally adaptive, adjustments to the function in one region of the input domain do not cause disruptive global changes \cite{refk-14}. This locality is particularly well-suited for modeling the nonstationary dynamics of EEG signals. Consequently, the network can learn to handle localized artifacts or abrupt ictal discharges without corrupting its response to normal background activity. Furthermore, this mathematical property also improves generalization. By learning a composition of simple, local functions, the model avoids overfitting to the unique global patterns of any single patient’s EEG, which in turn leads to stronger performance on unseen data \cite{refk-13, refn1, refn2}.
	
	Finally, this structural design directly contributes to improved parameter and data efficiency. Foundational research demonstrates that KANs frequently achieve higher accuracy with significantly fewer parameters than traditional MLPs, exhibiting more favorable scaling laws \cite{refk-14, refk-10}. Such efficiency is critical for seizure detection, where labeled data is often scarce and highly imbalanced. A model requiring fewer parameters is inherently less prone to overfitting on small datasets and can extract more meaningful representations from limited examples. Moreover, this compact architecture reduces the overall computational load, making KANs highly suitable for deployment on resource‑limited hardware, such as wearable EEG sensors or bedside monitoring units, where memory and power are in very limited supply \cite{refk-13}.

	\subsection{Review of Key KAN-Based Architectures for Seizure Detection}
	\label{subsec:kan_review}
	
	The emerging Kolmogorov-Arnold Network paradigm is rapidly moving from theoretical formulation to practical application in EEG-based seizure detection and prediction. Recent studies have explored various ways to integrate KANs,ranging from standalone models to hybrid architectures,to exploit their learnable edge functions.
	
	Ben~Attallah~et~al.~\cite{refk-19} demonstrated a prominent approach for integrating KANs into graph-based models to better capture spatial relationships between brain regions, introducing SeizureNet-KAN. This framework embeds KAN layers within a Graph Convolutional Network (GCN). Instead of using standard fixed activations like ReLU, the network uses learnable B-spline functions on its edges. This allows the model to capture the highly non-linear dynamics of pre-ictal (pre-seizure) EEG without requiring exceptionally deep layers. To handle data scarcity, the authors paired this compact architecture with a self-supervised learning framework, allowing the model to learn from unlabelled data. The resulting model is highly parameter-efficient while achieving state-of-the-art accuracy, demonstrating that KANs can effectively streamline complex spatial-temporal models.
	
	Hasan~et~al.~\cite{refk-20} highlighted the distinct challenges of deploying KANs as standalone temporal models by conducting a direct comparison between a pure KAN model and an LSTM network for seizure prediction. While the KAN utilized its learnable B-spline edges to model non-linear patterns, it achieved lower accuracy than the LSTM. The authors attributed this gap to a lack of inherent temporal memory. Specifically, the spline-based transformations operate on individual time windows without the explicit long-range recurrence mechanisms found in LSTMs. Despite this, the standalone KAN demonstrated a massive advantage in computational efficiency, converging significantly faster than the LSTM. This highlights an important design insight that pure KANs are fast and adaptable, yet they often need complementary recurrent or attention‑based mechanisms to fully capture temporal dependencies.
	
	Herboso~Contreras~et~al.~\cite{refk-2} addressed the need for highly deployable models by developing KAN-EEG, a shallow, end-to-end framework designed explicitly to replace standard MLP backbones. By utilizing just two KAN layers and significantly reducing the hidden parameter sizes, this model directly targets the demands of resource-constrained environments. Crucially, despite being trained on a fraction of the data required by comparable Transformer or ConvLSTM models, the smallest KAN-EEG variant demonstrated superior out-of-sample performance across entirely different, unseen datasets. This behavior directly confirms that the compact, localized nature of KANs makes them less prone to overfitting, offering a concrete solution to the cross-patient and cross-center generalization challenge while reducing memory footprints to levels suitable for edge computing devices.
	
	Peng~et~al.~\cite{refk-22} proved that KANs are highly effective when used selectively at the decision-making stage of larger hybrid models, introducing KAN-MemViT. This architecture uses a Memory-Cached Vision Transformer to extract complex temporal and spatial EEG features, but replaces the traditional MLP classification head with a two-layer KAN. The KAN acts as a highly flexible, non-linear decision boundary that better separates complex pre-seizure patterns. By using a KAN at the final output stage, the model not only reduces the false-alarm rate, which is a vital metric for clinical usability, but also enhances interpretability, as the learned edge functions governing the final classification can be explicitly visualized and analyzed.

	All these reviewed KAN‑based architectures are briefly summarised in Table~\ref{tab:kan_review}.
	
	\begin{table}[!htbp]
        \centering
        
        \begin{threeparttable}
            \caption{Summary of KAN‑Based Architectures for EEG Seizure Detection.}
            \label{tab:kan_review}
            \tiny
            \renewcommand{\arraystretch}{1.3}
            \setlength{\tabcolsep}{3pt}
            
            \begin{tabular}{
                    p{1.4cm} 
                    p{1.3cm} 
                    p{3.6cm} 
                    p{4.2cm} 
                    p{4.2cm} 
                }
                \toprule
                \textbf{Study} & \textbf{Dataset} & \makecell[l]{\textbf{Preprocessing /}\\ \textbf{Feature Extraction (Selection)}} & \textbf{Classification Architecture} & \textbf{Key Results (\%)} \\
                \midrule
                
                \makecell[l]{Ben~Attallah \\et al.\\ 2026 \cite{refk-19}} & 
                CHB-MIT & 
                EEG to graph via PLV;\newline dynamic thresholding;\newline statistical node features & 
                fusion-based GCN–KAN(KAGCN)\newline + hybrid SSL; KAN‑based decoder & 
                \makecell[l]{ACC - PRE - REC - AUC - F1 \\
                    97.68 - 97.72 - 97.53 - 99.28 - 97.59} \\ \\
                
                Hasan et al.\newline 2025 \cite{refk-20} & 
                CHB-MIT & 
                1-s sliding window (50\% overlap);\newline PCA; SMOTE & 
                LSTM: LSTM(64, 32, 16) $\rightarrow$ BatchNorm \newline $\rightarrow$ Dropout $\rightarrow$ Dense(128, 128) \newline $\rightarrow$ Dropout $\rightarrow$ Dense(1)\newline
                KAN: KAN(64, 32, 16) $\rightarrow$ BatchNorm $\rightarrow$ \newline Dropout $\rightarrow$ Dense(128) $\rightarrow$ BatchNorm$\rightarrow$ \newline Dropout $\rightarrow$ Dense(1) & 
                \makecell[l]{ACC - PRE - REC - F1 \\
                    80.15 - 64.00 - 80.00 - 71.00 \tnote{$\dagger$} \\
                    95.37 - 96.61 - 95.80 - 80.00 \tnote{$\ddagger$}} \\ \\
                
                Herboso~Contreras\newline et al.~2025 \cite{refk-2} & 
                TUH\newline (train/val),\newline EPILEPSIAE,\newline RPAH & 
                12-s window segmentation; ICA;\newline STFT (1 s window, 50\% overlap) & 
                KAN(764-256, 32-32, 32-16; Spline-based) \newline $\rightarrow$ Dense(1) $\rightarrow$ Sigmoid & 
                \makecell[l]{AUROC (TUH - EPILEPSIAE - RPAH) \\
                    88.7 - 73.0 - 83.0 \tnote{$\mathsection$} \\
                    86.9 - 78.0 - 85.0 \tnote{$\mathparagraph$} \\
                    88.9 - 55.0 - 60.0 \tnote{$\parallel$}} \\ \\
                
                Peng et al.\newline 2026 \cite{refk-22} & 
                CHB-MIT & 
                Band-pass filtering; Pre-emphasis;\newline 20-40 ms sliding window;\newline Hamming window; FFT ;\newline Mel filter bank; Log energy; DCT \newline MFCC selection & 
                SI-block (CNN + RFAConv) $\rightarrow$ \newline MCE Transformer $\rightarrow$ HF Module (CNN) \newline $\rightarrow$ KAN(2-layer) & 
                \makecell[l]{SEN - FPR - AUC \\
                    97.86 - 0.028/h - 95.10 \tnote{$\sharp$} \\
                    98.57 - 0.026/h - 0.891 \tnote{$\star$}} \\ 
                
                \bottomrule
            \end{tabular}
        
            \begin{tablenotes}\footnotesize
                \item[$\dagger$] Results obtained using KAN architecture.
                \item[$\ddagger$] Results obtained using LSTM architecture.
                \item[$\mathsection$] Results obtained using Compact(32-16) KAN architecture.
                \item[$\mathparagraph$] Results obtained using Compact(32-32) KAN architecture.
                \item[$\parallel$] Results obtained using Large KAN architecture.
                \item[$\sharp$] Results evaluated under patient-specific scenario.
                \item[$\star$] Results evaluated under cross-patient scenario.
            \end{tablenotes}
        \end{threeparttable}
        
    \end{table}
	
	\section{Future Directions for KAN-Based Seizure Detection}
	\label{sec:future}
	
	Although Kolmogorov--Arnold Networks have recently emerged as a promising paradigm for EEG-based seizure detection, their translation into clinically reliable systems remains at an early stage. To bridge the gap between algorithmic innovation and real-world deployment, future research must address several unresolved challenges related to adaptability, generalization, robustness, interpretability, and clinical integration.
	
	A major limitation of current seizure detection systems is their limited ability to adapt to the longitudinal nature of EEG monitoring. In real clinical settings, EEG signals are highly non-stationary. Slow background dynamics are often interrupted by abrupt ictal transitions, and a patient's baseline brain activity may evolve over time due to medication changes, sleep cycles, aging, or disease progression. Conventional deep learning models are poorly suited to this setting because updating them with new patient data often leads to catastrophic forgetting of previously learned patterns. In contrast, KANs provide a potentially valuable alternative due to their locally adjustable spline-based transformations. Since local changes in the input space do not necessarily distort the entire learned function, future work should explore continual learning frameworks that explicitly exploit this localized plasticity. In particular, adaptive grid refinement mechanisms that increase spline resolution only in clinically relevant regions may allow deployed KAN models to update incrementally in real time while preserving previously acquired knowledge. Such capability would be essential for personalized and lifelong seizure monitoring.
	
	Another important direction is improving generalization across patients, recording conditions, and acquisition protocols. Variations in electrode montage, channel count, hardware characteristics, and patient-specific neurophysiology substantially affect EEG morphology, making cross-subject generalization difficult. Future KAN-based systems should therefore be evaluated using patient-independent protocols such as leave-one-subject-out validation and cross-dataset benchmarking. In addition, domain adaptation and transfer learning strategies should be integrated to reduce patient-specific bias and improve robustness across heterogeneous data sources. Montage-invariant or permutation-aware KAN architectures, potentially combined with graph-based encoders, may further enable flexible handling of variable channel configurations without requiring complete retraining. Federated learning also represents a promising avenue for multi-center model development while preserving data privacy.
	
	Robustness to noise and artifacts is equally critical for practical deployment. Long-term and ambulatory EEG recordings often contain substantial contamination from ocular activity, muscle movement, electrode drift, and environmental interference. Although the flexible functional mappings of KANs may provide some inherent resilience, this assumption has not yet been systematically validated. Future studies should therefore assess KAN-based models under realistic artifact conditions and develop artifact-aware training and evaluation protocols to ensure reliability in uncontrolled clinical environments.
	
	Beyond classification accuracy, clinical utility depends heavily on minimizing false alarms. High false positive rates reduce clinician trust and increase workload, which can severely limit adoption even when sensitivity is high. For this reason, future KAN-based seizure detection systems should incorporate calibrated uncertainty estimation and confidence-aware decision mechanisms. These additions could help distinguish ambiguous events from reliable detections and support safer deployment in continuous monitoring settings.
	
	Interpretability also remains a key requirement for clinical acceptance. One of the most attractive features of KANs is their potential to represent nonlinear relationships in a more transparent and analyzable form than conventional black-box architectures. Future work should investigate how the symbolic or functional representations learned by KANs can be reviewed and validated by clinicians. Establishing clinician-in-the-loop workflows may allow KANs to serve not only as predictive tools but also as platforms for hypothesis generation and biomarker discovery.
	
	In addition, the limited availability of labeled seizure data highlights the importance of exploiting large volumes of unlabeled EEG recordings. Self-supervised learning strategies, such as masked segment reconstruction, contrastive learning, or future-window forecasting, may allow KAN-based architectures to learn robust latent transformations before fine-tuning on seizure labels. This direction may be especially beneficial for rare-event settings and low-seizure-frequency datasets, where annotation is costly and class imbalance is severe.
	
	Finally, future seizure detection systems are likely to operate within broader adaptive or closed-loop clinical frameworks, where model outputs may guide neurostimulation, alert generation, or therapeutic intervention. Integrating KAN-based models into such systems will require not only high accuracy, but also stable latency, calibrated prediction horizons, continuous adaptation, and rigorous safety validation. If these challenges are successfully addressed, KAN-based architectures may evolve from promising mathematical models into accurate, interpretable, and deployable tools for personalized seizure detection and prediction, thereby helping to bridge the gap between advanced neuro-AI research and real-world clinical practice.

	\section{Conclusion}
	\label{sec:conclusion}
	
	The automated detection of epileptic seizures via EEG remains a key challenge in computational neurology. As discussed in this review, the transition from heuristic, feature‑based machine learning approaches to deep learning paradigms has led to substantial improvements in classification accuracy. However, the clinical deployment of these standard deep architectures has been heavily bottlenecked by inherent challenges: the black-box opacity of standard activation functions, massive computational overhead, and extreme vulnerability to the non-stationarity of patient-specific EEG morphologies. 
	
	The recent emergence of Kolmogorov-Arnold Networks represents a highly promising paradigm shift specifically equipped to overcome these challenges. By embedding learnable functions on the network edges rather than relying on static nodal activations, KANs fundamentally align neural network topology with the localized, dynamic nature of neurophysiological signals. They provide a unique combination of high predictive accuracy, parameter efficiency, and strong mathematical interpretability. As the field moves toward integrating hybrid KAN architectures and exploring symbolic regression for enhanced clinical transparency, we anticipate a transformative shift from purely data‑driven black‑box models to interpretable, physics‑informed digital biomarkers. Ultimately, embracing these advanced mathematical frameworks will be critical to crossing the translational gap between algorithmic research and real-world, life-saving epilepsy care.
	
	\bibliographystyle{unsrtnat}
	\bibliography{bibliography}
	
\end{document}